\documentclass{article}

\usepackage{times}
\usepackage{graphicx} % more modern
\usepackage{subfigure} 

\usepackage{natbib}

\usepackage{algorithm}
\usepackage{algorithmic}

\usepackage{hyperref}
\usepackage{epsfig}
\usepackage{amsmath}
\usepackage{amssymb}
\usepackage{subfigure}
\usepackage{enumerate}
\usepackage{hyperref}
\usepackage[applemac]{inputenc}
\usepackage{pst-sigsys}
\usepackage{comment}
\usepackage{url}
\usepackage{hyperref}
\usepackage{breakurl}
\usepackage{subfigure}
\usepackage[accepted]{icml2014}
             
\newtheorem{theorem}{Theorem}[section]

\newtheorem{corollary}[theorem]{Corollary}

\newtheorem{proposition}[theorem]{Proposition}

\newcommand{\R}{{\mathbb{R}}}
\newcommand{\C}{{\mathbb{C}}}

\newcommand{\cQ}{{\mathcal Q}}
\newcommand{\cF}{{\mathcal F}}
\newcommand{\cJ}{{\mathcal J}}
\newcommand{\cS}{{\mathcal S}}
\newcommand{\cC}{{\mathcal C}}
\newcommand {\argmin}{arg\min\limits}
\newcommand\restr[2]{{% we make the whole thing an ordinary symbol
  \left.\kern-\nulldelimiterspace % automatically resize the bar with \right
  #1 % the function
  \vphantom{\big|} % pretend it's a little taller at normal size
  \right|_{#2} % this is the delimiter
  }}

\icmltitlerunning{Signal recovery from Pooling Representations}

\begin{document} 

\twocolumn[
\icmltitle{Signal recovery from Pooling Representations}

% It is OKAY to include author information, even for blind
% submissions: the style file will automatically remove it for you
% unless you've provided the [accepted] option to the icml2014
% package.
\icmlauthor{Joan Bruna}{bruna@cims.nyu.edu}
\icmladdress{Courant Institute of Mathematical Sciences,
           715 Broadway New York NY 10003 USA}
\icmlauthor{Arthur Szlam}{aszlam@ccny.cuny.edu}
\icmlauthor{Yann LeCun}{yann@cs.nyu.edu}
\icmladdress{Courant Institute of Mathematical Sciences,
           715 Broadway New York NY 10003 USA}

% You may provide any keywords that you 
% find helpful for describing your paper; these are used to populate 
% the "keywords" metadata in the PDF but will not be shown in the document
\icmlkeywords{boring formatting information, machine learning, ICML}

\vskip 0.3in
]
\begin{abstract} 
%Pooling operators construct non-linear representations by cascading a redundant linear transform, followed by  a point-wise nonlinearity and a local aggregation, typically implemented with a $\ell_p$ norm.  Their efficiency in recognition architectures is based  on their ability to locally contract the input space,  but also on their capacity to retain as much stable information  as possible.
%We address this latter question by computing the upper and 
In this work we compute
lower Lipschitz bounds of $\ell_p$ pooling operators for $p=1, 2, \infty$
as well as $\ell_p$ pooling operators preceded by half-rectification layers.  These give
sufficient conditions for the design of invertible neural network layers.
Numerical experiments on MNIST and image patches confirm that
pooling layers can be inverted with phase recovery algorithms. Moreover,
the regularity of the inverse pooling, controlled by the lower Lipschitz constant,
 is empirically verified with a nearest neighbor regression.
\end{abstract} 
\section{Introduction}

A standard architecture for deep feedforward networks  consists of a number of stacked modules,  each of which consists of a linear mapping, followed by an elementwise nonlinearity,  followed by a pooling operation.  
%Deep neural networks have proven to be generic and highly efficient signal representations for a variety of recognition tasks. 
%They construct local invariants by cascading linear decompositions, element-wise nonlinearities and pooling, which computes a local statistic over a collection of measurements.
Critical to the 
success of this architecture in recognition problems is its capacity for preserving discriminative signal information while being invariant to nuisance deformations.   The recent works \cite{stephane,bruna_pami} study the role of the pooling operator in building invariance.  In this work, we will study a network's capacity for preserving information.  Specifically, we will study the invertibility of modules with a linear mapping,
the half rectification nonlinearity, and $\ell_p$ pooling, for $p\in \{1,2,\infty\}$.  We will discuss recent work in the case $p=2$, and connections with the phase recovery problem of \cite{phaselift,GSalgo,1206.0102}.

%
%%A standard architecture for deep feedforward networks  consists of a number of stacked modules, 
%%each of which consists of a linear mapping, followed by an elementwise nonlinearity, 
%%followed by a pooling operation.  
%Deep neural networks have proven to be generic and highly
%efficient signal representations for a variety of recognition tasks. 
%They construct local invariants by cascading linear decompositions,
%element-wise nonlinearities and pooling, which computes a local statistic 
%over a collection of measurements.
%This basic building block has the capacity to produce local invariance, 
%but its efficiency also depends
%upon its ability to preserve signal information.
%In this work we will explore the invertibility of these layers,
% using the half rectification nonlinearity and $\ell_p$ pooling, for $p\in \{1,2,\infty\}$.  
% % discussing recent work in the case $p=2$, and especially connections with the phase recovery problem of \cite{phaselift,GSalgo,1206.0102}.

\subsection{$\ell_p$ pooling}
The purpose of the pooling layer in each module is to give invariance to the system, 
perhaps at the expense of resolution.  This is done via a summary statistic over the 
outputs of groups of nodes.  In the trained system, the columns of the weight matrix 
 corresponding to nodes grouped together often exhibit similar characteristics, and 
 code for perturbations of a template \cite{koray-cvpr-09,hyvarinen_topo}.  

The summary statistic in $\ell_p$ pooling is the $\ell_p$ norm of the inputs 
into the pool. That is, if nodes $x_{I_i},...,x_{I_l}$ are in a pool, the output of the pool is 
\[\left(|x_{I_i}|^p+...+|x_{I_l}|^p\right)^{1/p},\]
where as usual, if $p\rightarrow \infty$, this is 
\[\max \left(|x_{I_i}|,...,|x_{I_l}|\right).\]
%Let us fix a single layer of the system, with weight matrix $W$ and nonlinearity $s$.  The $i$th output of the layer with $\ell_p$ pooling and input $x$ is \[\left(\sum_{j \in P_i} s(Wx)_j^p\right)^{1/p}\]
If the outputs of the nonlinearity are nonnegative (as for the half rectification function), 
then $p=1$ corresponds to average pooling, and the case $p=\infty$ is max pooling.
% and the case $p=2$ has been used to promote group sparsity.% \cite{}.

\subsection{Phase reconstruction}

Given  $x\in \R^n$, a classical problem in signal processing is to recover $x$ 
from the absolute values of its (1 or 2 dimensional) Fourier coefficients, perhaps 
subject to some additional constraints on $x$; this problem arises in speech 
generation and X-ray imaging \cite{yonina}.  Unfortunately, the problem is not well 
posed- the absolute values of the Fourier coefficients do not nearly specify $x$.  
For example, the absolute value of the Fourier transform is translation invariant. %to rotations
%of the coordinates.  
 It can be shown (and we discuss this below) that the 
 absolute value of the inner products between $x$ and any basis of $\R^n$ 
 are not enough to uniquely specify an arbitrary $x$; the situation is worse for $\C^n$. 
  On the other hand, recent works have shown that by taking a redundant enough dictionary, 
  the situation is different, and $x$ can be recovered from the modulus 
  of its inner products with the dictionary \cite{Balan2006,phaselift,1206.0102}.

Suppose for a moment that there is no elementwise nonlinearity in our 
feedforward module, and only a linear mapping followed by a pooling.
Then with a slightly generalized notion of phase, where  the modulus 
is the $\ell_p$ norm of the pool, and the phase is the $\ell_p$ unit vector 
specifying the ``direction'' of the inner products in the pool, the phase 
recovery problem above asks if the module loses any information.  The $\ell_2$ case has been recently studied in \cite{1305.6226}

\subsection{What vs. Where} 
%need more here, and some citations!!!
If the columns of the weight matrix in a pool correspond to related 
features, it can be reasonable to see 
the entire pool as a ``what''.   That is, the modulus of the pool indicates the 
relative presence of a grouping of (sub)features into a template, and the phase 
of the pool describes the relative arrangement of the subfeatures, describing 
``where'' the template is, or more generally, describing the ``pose'' of the template.

From this viewpoint, phase reconstruction results make rigorous the 
notion that given enough redundant versions of ``what'' and throwing away the ``where'', we can still recover the ``where''.

\subsection{Contributions of this work}
%this section needs work
In this work we  give conditions so that a module consisting of a linear mapping, perhaps followed by a half rectification, followed by an $\ell_p$ pooling preserves the information in its input.  We extend the  $\ell_2$ results of \cite{1305.6226,1308.4718} in several ways: we consider the  $\ell_p$ case, 
take into account the half rectification nonlinearity, and we make the results quantitative in the sense that we give bounds on the lower Lipschitz constants of the modules.  This gives a measure of the {\it stability} of the inversion, which is especially important in a multi-layer system.  
Using our bounds, we prove that redundant enough random modules with $\ell_1$ or $\ell_{\infty}$ pooling are invertible.   

We also show the results of numerical experiments designed to explore the gaps in our results and the results in the literature.
We note that the alternating minimization method of \citep{GSalgo} 
can be used essentially unchanged for  the  $\ell_p$ case, with or without 
rectification, and show experiments giving evidence that recovery is roughly 
equally possible for $\ell_1$, $\ell_2$, and $\ell_\infty$ using this 
algorithm; and that half rectification before pooling can make recovery easier. Furthermore, we show that with a trained  initialization, 
the alternating method compares favorably with the state of the art recovery methods (for $\ell_2$ with no rectification) in \cite{1206.0102,phaselift}, which suggests that the above observations are not an artifcact  of the alternating method.

\section{Injectivity and Lipschitz stability of Pooling Operators}
\label{sect2}

This section studies necessary and sufficient conditions guaranteeing 
that pooling representations are invertible. It also computes 
 upper and lower Lipschitz bounds, which are tight under certain configurations.

Let us first introduce the notation used throughout the paper. 
Let $\cF=\{f_1,\dots,f_M\}$ be a real frame of $\mathbb{R}^N$, with $M>N$. 
The frame $\cF$ is organized into $K$ disjoint blocks $\cF_k=\{f_j\}_{j \in I_k}$, $k=1\dots K$, 
such that $I_k \cap I_{k'} = \emptyset$ and $\bigcup_k I_k = \{1\dots M\}$. 
For simplicity, we shall assume that all the pools have equal size $|I_k|=L$.

The $\ell_p$ pooling operator $P_p(x)$ is defined as the mapping
\begin{equation}
\label{lppool}
x \mapsto P_p(x)=\{ \| \cF_k^T x \|_p \,\,,\, k=1\dots K\}~.
\end{equation}

A related representation, which has gained popularity in recent deep learning
architectures, introduces a point-wise thresholding before computing
the $\ell_p$ norm. If $\alpha \in \mathbb{R}^M$ is a fixed threshold vector, and
$(\rho_\alpha(x))_i = \max( 0, x_i- \alpha_i)$, then the $\ell_p$ \emph{rectified pooling} 
operator $R_p(x) $ is defined as 
\begin{equation}
\label{lprect}
x \mapsto R_p(x)=\{ \| \rho_{\alpha_k} (\cF_k^T x) \|_p \,\,,\, k=1\dots K\}~, 
\end{equation}
where $\alpha_k$ contains the coordinates $I_k$ of $\alpha$.

We shall measure the stability of the inverse pooling 
with the Euclidean distance in the representation space.
Given a distance $d(x,x')$ in the input space,
the Lipschitz bounds of a given operator $\Phi(x)$ are defined as 
the constants $0\leq A\leq B$ such that 
$$\forall\,\,x\,,\,x'~,~ A d(x,x') \leq \| \Phi(x) - \Phi(x') \|_2 \leq B d(x,x')~.$$

In the remainder of the paper, given a frame $\cF$, we denote respectively by 
$\lambda_{-}(\cF)$ and $\lambda_{+}(\cF)$ its lower and upper frame bounds. 
If $\cF$ has $M$ vectors and $\Omega \subset \{1,\dots,M\}$,
we denote $\cF_\Omega$ the frame obtained by keeping the vectors 
indexed in $\Omega$.
Finally, we denote ${\Omega}^c$ the complement of $\Omega$.

\subsection{Absolute value and Thresholding nonlinearities}
\label{abs_and_rect}

In order to study the injectivity of pooling representations, 
we first focus on the properties of the operators defined 
by the point-wise nonlinearities.

The properties of the \emph{phaseless} mapping 
\begin{equation}
\label{modudef}
x \mapsto M(x)=\{ | \langle x, f_i \rangle |\,,\, i=1\dots m \}~,~ x \in \mathbb{R}^n~,
\end{equation} 
have been extensively studied in the literature
\cite{Balan2006, 1308.4718}, in part motivated by 
applications to speech processing \cite{speech_phaseless} 
or X-ray crystallography \cite{yonina}. 
It is shown in \cite{Balan2006} that if $m > 2n-1$ then 
it is possible to recover $x$ from $M(x)$, up to a global
sign change. 
In particular, \cite{1308.4718} recently characterized
the stability of the phaseless operator, that is 
summarized in the following proposition:

\begin{proposition}[\cite{1308.4718}, Theorem 4.3]
\label{balanprop}
Let $\cF=(f_1,\dots,f_M)$ with $f_i \in \mathbb{R}^N$ 
and \\ $d(x,x') = \min(\| x - x' \|, \| x + x'\|)$.
The mapping $M(x)=\{ |\langle x, f_i \rangle |\}_{i\leq m}$ 
satisfies 
\begin{equation}
\label{mod_lipsch}
\forall\,x,x' \in \mathbb{R}^n~,~ A_{\cF}\,d(x,x')  \leq \| M(x) - M(x') \| \leq B_{\cF} \,d(x,x') ~,
\end{equation}
where 
\begin{eqnarray}
\label{mod_lipsch_conds}
A_{\cF}&=& \min_{\Omega \subset \{1\dots M\}} \sqrt{ \lambda_{-}^2( \cF_\Omega) + \lambda_{-}^2 (\cF_{{\Omega}^c})}~, \\
B_{\cF}&=& \lambda_{+}(\cF)~.
\end{eqnarray}
In particular, $M(x)$ is injective if and only if
for any subset $\Omega \subseteq \{1,\dots,M\}$, either
$\cF_\Omega$ or 
$\cF_{\Omega^c}$ is an invertible frame.
\end{proposition}
A frame $\cF$ satisfying the previous condition 
is said to be \emph{phase retrievable}.

We now turn our attention to the half-rectification operator, 
defined as 
\begin{equation}
\label{rectdef}
M_{\alpha}(x) = \rho_\alpha(\cF^T x)~.
\end{equation}
For that purpose, let us introduce some extra notation. 
Given a frame $\cF=\{f_1,\dots,f_M\}$, a subset $\Omega \subset \{1 \dots M\}$ 
is \emph{admissible} if 
\begin{equation}
\label{rect_partition}
\bigcap_{i \in \Omega} \{ x \,;\, \langle x, f_i \rangle > \alpha_i \}  \cap \bigcap_{i \notin \Omega} \{ x \,;\, \langle x, f_i \rangle < \alpha_i \} \neq \emptyset~.
\end{equation}
We denote by $\overline{\Omega}$ the collection of all admissible sets,
and $V_\Omega$ the vector space generated by $\Omega$.
The following proposition, proved in Section \ref{proofs}, 
gives a necessary and sufficient condition for the
injectivity of the half-rectification.
\begin{proposition}
\label{halfrect_inj}
Let $A_0= \min_{\Omega \in \overline{\Omega}} \lambda_{-}(\restr{\cF_\Omega}{V_\Omega})$.
Then the half-rectification operator $M_{\alpha}(x) = \rho_\alpha(\cF^T x)$ 
is injective if and only if $A_0>0$. Moreover,
it satisfies 
\begin{equation}
\label{halfrect_lips}
\forall\,x,x'~,~ A_0 \| x - x' \| \leq \| M_{\alpha}(x) - M_{\alpha} (x') \| \leq B_0 \|x-x' \|~,
\end{equation}
with  
$B_0=\max_{\Omega \in \overline{\Omega}} \lambda_{+}(\cF_\Omega) \leq \lambda_{+}(\cF)$.
\end{proposition}

The half-rectification has the ability to recover 
the input signal, without the global sign ambiguity. 
The ability to reconstruct from $M_\alpha$ is thus
controlled by the rank of any matrix $\cF_\Omega$ whose
columns are taken from a subset belonging to $\overline{\Omega}$.
In particular, if $\alpha \equiv 0$,
since $\Omega \in \overline{\Omega} \Rightarrow \Omega^c \in \overline{\Omega}$, 
it results that $m\geq 2n$ is necessary in order 
to have $A_0>0$.

The rectified linear operator
 creates a partition of the input space into 
 polytopes, defined by (\ref{rect_partition}), 
 and computes a linear operator 
on each of these regions.
A given input $x$ activates a set $\Omega_x \in \overline{\Omega}$, 
encoded by the sign of the linear measurements
$\langle x, f_i \rangle - \alpha_i$. 

As opposed to the absolute value 
operator, the inverse of $M_\alpha$, whenever 
it exists, can be computed directly by locally inverting
a linear operator. Indeed, 
the coordinates of $M_\alpha(x)$ satisfying
$M_\alpha(x)_j > \alpha_j$ form a set $s(x)$,
which defines a linear 
model $\cF_{s(x)}$ which is invertible if $A_0 > 0$.
%However, this property is not useful 
%as soon as the rectification is cascaded, since 
%one is no longer able to identify the linear model
%to be inverted by directly inspecting the coordinates.

%(The condition can be particularized on a region of the space)

In order to compare the stability of the half-rectification
versus the full rectification, one can modify $M_{\alpha}$
so that it maps $x$ and $-x$ to the same point. If 
one considers 
$$\widetilde{M}_{\alpha}(x) = \left \{ 
\begin{array}{rl}
M_{\alpha}(x) & ~\mbox{if } \lambda_{-}(\cF_{s(x)}) > \lambda_{-}(\cF_{{s(x)}^c}) ~,\\
M_{-\alpha}(-x) & ~\mbox{otherwise} ~.
\end{array} \right.
$$
then $\tilde{M}_\alpha$ satisfies the following:
\begin{corollary}
\label{rectif_corollary}
\begin{equation}
\forall\,x,x' \in \mathbb{R}^n~,~ \widetilde{A}\,d(x,x')  \leq \| \widetilde{M}_\alpha(x) - \widetilde{M}_\alpha(x') \| \leq \widetilde{B} \,d(x,x') ~,
\end{equation}
with
\begin{eqnarray}
\label{rect_lipsch_conds}
\widetilde{A}&=& \min_{\Omega \subset \overline{\Omega}} \max(\lambda_{-}^2( \cF_\Omega) ,\, \lambda_{-}^2 (\cF_{\Omega^c})) ~,  \\
\widetilde{B}&=& \max_{\Omega \subset \overline{\Omega}} \lambda_{+}(\cF_\Omega) \leq \lambda_{+}(\cF)~,
\end{eqnarray}
and $d(x,x')=\min(x-x',x+x')$, so 
$\widetilde{A} \geq 2^{-1/2} A $ and %\sqrt{ \lambda_{-}^2( \cF_S) + \lambda_{-}^2 (\cF_{\overline{S}})}$ and
$\widetilde{B} \leq B$.
In particular, if $M$ is invertible, so is $\tilde{M}_\alpha$.
\end{corollary}

It results that the bi-Lipschitz bounds of the half-rectification
operator, 
when considered in under the equivalence $x \sim -x$, 
 are controlled by the bounds of the 
absolute value operator, up to a factor $2^{-1/2}$. However, 
the lower Lipschitz bound (\ref{rect_lipsch_conds}) 
consists in a minimum taken over 
a much smaller family of elements than 
in (\ref{mod_lipsch_conds}).

\subsection{$\ell_p$ Pooling}

We give bi-Lipschitz constants of the 
$\ell_p$ Pooling and $\ell_p$ rectified Pooling 
operators for $p=1,2,\infty$. 
%From now on, it is assumed that
%the vectors within each block $\cF_k = \cF_{I_k}$ are 
%orthogonal.

From its definition, it follows that 
 pooling operators $P_p$ and $R_p$ can be expressed 
 respectively as a function of phaseless and half-rectified 
 operators, which implies that for the pooling to be
invertible, it is necessary that the absolute value and rectified
operators are invertible too. Naturally, the 
converse is not true. 

\subsubsection{$\ell_2$ pooling}

The invertibility conditions of the $\ell_2$ pooling representation 
have been recently studied in \cite{1305.6226}, 
where the authors obtain necessary and sufficient
conditions on the frame $\cF$.
We shall now generalize those results, and
derive bi-Lipschiz bounds.

Let us define
\begin{equation}
\label{q2family}
\cQ_2 = \left \{  ( U_k \cF_k)_{k\leq K} \,;\,\forall \,k \leq K~,~ U_k \in \R^{d\times d}\,,\, U_k^T\, U_k = {\bf Id} \right\} ~.
\end{equation}
$\cQ_2$ thus contains all the orthogonal 
bases of each subspace $\cF_k$.

The following proposition, proved in section \ref{proofs},
computes upper and lower bounds of the
Lipschitz constants of $P_2$. %and $R_p$.

\begin{proposition}
\label{p2lips}
%Let $p=\{1,2,\infty\}$. 
The $\ell_2$ pooling operator $P_2$ 
satisfies
\begin{equation}
\label{p2lips_eq}
\forall~x,\,x'; ~,~A_2 d(x,x') \leq \| P_2(x) - P_2(x') \| \leq B_2 d(x,x') ~,
\end{equation}
where 
\begin{eqnarray}
\label{p2_lips_conds}
A_2 &=& \min_{\cF' \in \cQ_2} \min_{\Omega \subset \{1\dots M\}} \sqrt{ \lambda_{-}^2( \cF'_\Omega) + \lambda_{-}^2 (\cF'_{{\Omega}^c})}~, \nonumber \\
B_2 &=& \lambda_{+}(\cF)~.
\end{eqnarray}
%with $\alpha_1=\sqrt{d}$, $\alpha_2=1$ and $\alpha_\infty=d$. 
%Moreover, the rectified $\ell_p$ pooling $R_p$ also satisfies (\ref{p2lips_eq}) 
%with the same constants.
\end{proposition}
This proposition thus generalizes the results from \cite{1305.6226}, 
since it shows that $A_2 >0$ not only controls when $P_2$ is invertible, 
but also controls the stability of the inverse mapping.

We also consider the rectified $\ell_2$ pooling case.
For simplicity, we shall concentrate in the case where the pools have 
dimension $d=2$. %and $\alpha=0$.
For that purpose, for each $x,x'$, we consider 
a modification of the families ${\cQ}_2$, by replacing 
each sub-frame $\cF_k$ by $\cF_{I_k \cap s(x) \cap s({x'})}$, 
that we denote $\widetilde{\cQ}_{2,x,x'}$. 
\begin{corollary}
\label{rectpoolcor}
Let $d=2$, and set $p(x,x')=s(x) \cup s({x'}) \backslash (s(x) \cap s({x'}))$. %and $\alpha=0$. 
Then the rectified $\ell_2$ pooling operator $R_2$ satisfies 
\begin{equation}
\label{p2lips_eqbis}
\forall~x,\,x'; ~,~\tilde{A}_2 d(x,x') \leq \| R_2(x) - R_2(x') \| \leq B_2 d(x,x') ~,
\end{equation}
where 
\begin{eqnarray}
\label{p2_lips_condsbis}
\tilde{A}_2 &=& \inf_{x,x'} \min_{\cF' \in \widetilde{\cQ}_{2,x,x'}}\min_{\Omega \subset s(x) \cap s({x'})} \Big( \lambda_{-}^2(\cF_{p(x,x')}) + \nonumber \\ 
&& \lambda_{-}^2( \cF'_\Omega) + \lambda_{-}^2 (\cF'_{{\Omega}^c}) \Big)^{1/2}~, \nonumber 
\end{eqnarray}
\end{corollary}

Proposition \ref{p2lips} and Corollary \ref{rectpoolcor}  
give a lower Lipschitz bound which gives sufficient guarantees
for the inversion of pooling representations. 
Corollary \ref{rectpoolcor} indicates that, in the case $d=2$, 
the lower Lipschitz bounds are sharper than the non-rectified case,
in consistency with the results of section \ref{abs_and_rect}. 
The general case $d>2$ remains an open issue.

\subsubsection{$\ell_\infty$ Pooling}
\label{sec:linfty}
We give in this section sufficient and necessary conditions 
such that the max-pooling operator $P_\infty$ is injective, 
and we compute a lower bound of its lower Lipschitz constant.

%define switches
Given $x \in \R^N$, we define the \emph{switches} 
$s(x)$ of $x$ as the $K$ vector of coordinates in each pool where the maximum 
is attained; that is, for each $k\in \{1,\dots, K\}$:
$$s(x)_k = \arg\max_{j \in I_k} | \langle x, f_j \rangle |, $$
and we denote by $\cS$ the set of all attained switches: $\cS = \{ s(x) \, ; \, x \in \R^N\}$.
This is a discrete subset of $\prod_k \{1,\dots,I_k\}$.
%define cones
Given $s \in \cS$, the set of input signals having $s(x)=s$ defines a 
linear cone $\cC_s \subset \R^N$:
$$\cC_s = \bigcap_{k\leq K} \bigcap_{j \in I_k} \{ x \,;\, | \langle x, f_{s_k} \rangle | \geq | \langle x, f_j \rangle | \}~,$$
and as a result the input space is divided into a collection of 
Voronoi cells defined from linear equations. Restricted to each cone 
$\cC_s$, the max-pooling operator computes the phaseless mapping $M(x)$ from equation \eqref{modudef} corresponding to $\cF_s = (f_{s_1},\dots,f_{s_K})$.

%For each cone $\cC \subset \R^n$, we define its projection $\Pp_{\cC}$ 
%as the mapping $x \mapsto \langle x, e^* \langle e^*$, where 
%$e^*$ is the closest element of the unit ball of $\R^n$ to the direction $x / \|x \|$. 

%define the angle between switches
Given vectors $u$ and $v$, as usual, set the angle $\theta(u,v)=\arccos \left( \frac{| \langle u, v \rangle |}{\| u \| \|v\|} \right).$
For each $s, s' \in \cS$ such that $\cC_s \cap \cC_{s'} = \emptyset$
and for each $\Omega \subset \{1\dots K\}$, we define 
$$\beta(s, s',\Omega) = \min_{ u \in \restr{\cF_s}{\Omega}(\cC_s) \,\, v \in \restr{\cF_{s'}}{\Omega}(\cC_{s'})} \theta(u,v).$$
This is a modified first principal angle between the subspaces $\restr{\cF_s}{\Omega}$ and $\restr{\cF_{s'}}{\Omega}$, 
where the infimum is taken only on the directions included in the respective cones.
Set $\Lambda_{s,s',\Omega}=\lambda_{-}(\restr{\cF}{\Omega})\cdot \sin(\beta(s,s',\Omega))$.
 
Given $s$, $s'$, we also define $\cJ(s,s') = \{ k \, ; \, s_k = s'_k \}$.
Recall $L$ is the size of each pool.    Set

\begin{eqnarray*}
%\label{mpool_bobo}
A(s,s') &=& \Big \{ \min_{\Omega \subseteq \cJ(s,s')} \lambda_{-}^2(\cF_\Omega) + \lambda_{-}^2(\cF_{\cJ - \Omega})  + \nonumber \\
& & \frac{1}{4L} \, \min_{\Omega \subseteq \cJ(s,s')^c} \Lambda_{s,s',\Omega}^2 + \Lambda_{s,s',\Omega^c}^2  \Big \}^{1/2}  ~. 
\end{eqnarray*}

The following proposition, proved in section \ref{proofs}, gives a lower Lipschitz bound of the max-pooling operator.
%Lipschitz proposition
\begin{proposition}
\label{mpooling_prop}
For all  $x$ and $x'$, the max-pooling operator $P_\infty$ satisfies
\begin{equation}
\label{cc3}
d(x,x') \left(\min_{s,s'} A(s,s') \right) \leq \| P_\infty(x) - P_\infty(x') \|,
\end{equation}
where $d(x,x')=\min(\|x-x'\|,\|x+x'\|)$.

\end{proposition}

Propostion \ref{mpooling_prop} shows that the lower Lipschitz bound 
is controlled by two different phenomena. The first one depends upon 
how the cones corresponding to disjoint switches are aligned, whereas the
second one depends on the internal incoherence of each 
frame $\cF_{\cJ(s,s')}$.  
One may ask how do these constants evolve in different asymptotic regimes. 
For example, if one lets the number of pools $K$ be fixed but increases the 
size of each pool by increasing $M$.  
In that case, the set of
possible switches $\cS$ increases, and each cone $\cC_s$ gets smaller. 
The bound corresponding to $\cC_s \cap \cC_{s'} \neq \emptyset$ 
decreases since the infimum is taken over a larger
family. However, as the cones $\cC_s$ become smaller, 
the likelihood that any pair $x\neq x'$ share the same switches decreases,
thus giving more importance to the case $\cC_s \cap \cC_{s'} = \emptyset$. 
Although the ratio $\frac{1}{L}$ decreases, the lower frame
bounds $\lambda_{-}(\cF_{\Omega})^2$, $\lambda_{-}(\cF_{\Omega^c})^2$ 
will in general increase linearly with $L$. The lower bound
will thus mainly be driven by the principal angles 
$\beta(s,s',\Omega)$. Although the minimum in (\ref{cc3}) 
is taken over a larger family, each angle is computed over a smaller
region of the space, suggesting that one can indeed increase 
the size of each pool without compromising the injectivity of the max-pooling.

Another asymptotic regime considers pools of fixed size $L$ and 
increases the number of pools $K$. In that case, the bound 
increases as long as the principal angles remain 
lower bounded. 

We also consider the stability of max-pooling with a half-rectification.
By redefining the switches $s(x)$ accordingly:
\begin{equation}
\label{switches_rect}
s(x)=\{ j \, ; \, \langle x, f_j \rangle + \alpha_j > \max(0, \langle x, f_{j'}\rangle + \alpha_{j'} \,; \,\forall~j' \in \mbox{pool}(j)\} ~,
\end{equation}
the following proposition, proved in section \ref{proofs}, computes a lower bound 
of the Lipschitz constant of $R_\infty$.
\begin{corollary}
\label{mpoolrectprop}
The rectified max-pooling operator $R_\infty$ satisfies
\begin{equation}
\label{cc34}
\forall\, x,x'~, \|x - x'\| \min_{s,s'} A(s,s') \leq \| R_\infty(x) - R_\infty(x') \|~,
\end{equation}
with 
\begin{equation*}
%\label{mpool_bobo}
A(s,s') = \Big \{ \lambda_{-}^2(\cF_{\cJ(s,s')}) + \frac{1}{4L} \Lambda_{s,s',\cJ(s,s')^c}^2  \Big \}^{1/2}~
\end{equation*}
defined using the cones $\cC_s$ obtained from (\ref{switches_rect}).
 \end{corollary}
%
%These results easily extend to the so-called Maxout operator \cite{maxout}, 
%defined as $x \mapsto MO(x)=\{ \max_{j \in I_k} \langle x, f_j \rangle \, ; \, k=1\dots K\}$. 
%By redefining the switches of $x$ as
%\begin{equation}
%\label{switches_maxout}
%s(x)=\{ j \, ; \, \langle x, f_j \rangle  > \max( \langle x, f_{j'}\rangle \,; \,\forall~j' \in \mbox{pool}(j)\} ~,
%\end{equation}
%the following corollary computes a Lower Lipschitz bound of $MO(x)$:
%\begin{corollary}
%\label{cormaxout}
%The Maxout operator $MO$ satisfies (\ref{cc34}) with $A(s,s')$ defined 
%using the switches (\ref{switches_maxout}).
%\end{corollary}

\subsubsection{$\ell_1$ Pooling and Max-Out}
\label{sec:l1}
Propostion \ref{mpooling_prop} can be used to 
obtain a bound of the lower Lipschitz constant 
of the $\ell_1$ pooling operator, 
as well as the Maxout operator \cite{maxout};
 see section \ref{sec:l1sup} in the supplementary material.

\subsection{Random Pooling Operators}
What is the minimum amount of redundancy needed to 
invert a pooling operator? As in previous works on compressed
sensing \cite{csensing} and phase recovery \cite{Balan2006}, 
one may address this question by studying random pooling 
operators. 
In this case, the lower Lipschitz bounds derived 
in previous sections can be shown to be positive with probability
$1$ given appropriate parameters $K$ and $L$.

The following proposition, proved in Appendix \ref{proofs}, analyzes the invertibility of a generic
 pooling operator constructed from random measurements.
We consider a frame $\cF$ where its $M$ columns are iid Gaussian vectors of $\R^N$. 
%\begin{proposition}
\begin{proposition}
\label{randomprop}
Let $\cF=(f_1,\dots,f_M)$ be a random frame of $\R^N$, organized
into $K$ disjoint pools of dimension $L$.
 %Then these statements hold with probability $1$:
 With probability $1$
%\begin{enumerate}
%\item 
$P_p$ is injective (modulo $x \sim -x$) if $K \geq 4N$ for $p=1,\infty$ 
and if $K \geq 2N-1$ for $p=2$.
%\item The Maxout operator $MO$ is injective if $K \geq 2N+1$.
%\end{enumerate}
\end{proposition}
The size of the pools $L$ does not 
influence the injectivity of random pooling, 
but it affects the stability of the inverse,
as shown in proposition \ref{mpooling_prop}.
The half-rectified case requires extra care,  since 
 the set of admissible switches $\overline{\Omega}$ might contain 
 frames with $M<N$ columns with non-zero probability, 
 and is not considered in the present work.
% 
% DISCUSS CASE WITH STRUCTURED INPUT. 
% X IS IN A UNION OF LOW-DIMENSIONAL SUBSPACES. 
% HOW DOES THE NUMBER OF MEASUREMENTS CHANGE?
% 
 
\section{Numerical Experiments}
\label{sect3}
Our main goal in this section is to experimentally compare the invertibility of $\ell_p$ pooling for  $p\in\{1,2,\infty\}$, with and without rectification. Unlike in the previous sections, we will not consider the Lipschitz bounds, as we do not know a good way to measure these experimentally.   Our experiments suggest that recovery is roughly the same difficulty for $p=1,2,\infty$, and that rectification makes recovery easier.

In the $\ell_2$ case without rectification, and with $d=2$, a growing body of works \cite{phaselift,1206.0102} describe how to invert the pooling operator.  This is often called phase recovery.  A problem for us is a lack of a standard algorithm when $p\neq 2$ or with rectification.  We will see that the simple alternating minimization algorithm of \cite{GSalgo} can be adapted to these situations.   However, alternating minimization with random initialization is known to be an inferior recovery algorithm for $p=2$, and so any conclusions we will draw about ease of recovery will be tainted, as we would be testing whether the algorithm is equally bad in the various situations, rather than if the problems are equally hard.    
We will show that in certain cases, a training set allows us to find a good initialization for the alternating minimization, leading to excellent recovery performance, and that in this setting, or the random setting, recovery via alternating minimization is roughly as succesful for each of the $p$, suggesting invertibility is equally hard for each $p$.  In the same way, we will see evidence that half rectification before pooling makes recovery easier.

\subsection{Recovery Algorithms}
\subsubsection{Alternating minimization}
A greedy method for recovering the phase from the modulus of complex measurements is given in \cite{GSalgo}; 
this method naturally extends to the case at hand.  As above, denote the frame $\{f_1,...,f_M\}=\cF$, let $\cF_k$ be
 the frame vectors in the $k$th block, and set $I_k$ to be the indices of the $k$th block.
Let $\cF^{(-1)}$ be the pseudoinverse of $\cF$; set $(P_p(x))_k=||\cF_k x||_p$.  Starting with an initial signal $x^0$,  update 
\begin{enumerate}
\item $y^{(n)}_{I_k}=(P_p(x))_k \frac{\cF_k x^{(n)}}{||\cF_k x^{(n)}||_p}$,  $k=1\dots K$,
\item $x^{(n+1)}=\cF^{(-1)}y^{(n)}$.
\end{enumerate}
This approach is not, as far as we know, guarantee to converge to the correct solution, even when  $P_p$  is invertible.  
However, in practice, if the inversion is easy enough, or if $x_0$ is close to the true solution, 
the method can work well. Moreover,  
%For us, the important thing about 
this algorithm can be run essentially unchanged for each $\ell_p$; for half rectification, we only use the nonegative entries in $y$ for reconstruction.

In the experiments below, we will use random, Gaussian i.i.d. $\cF$, but also we will use the outputs 
of dictionary learning with block sparsity.  The $\cF$ generated this way is not really a frame, 
as the condition number of a trained dictionary on real data is often very high.  
In this case, we will renormalize each data point to have norm $1$, and modify the update $x^{(n+1)}=\cF^{(-1)}y^{(n)}$ to 
\vspace{-2.5mm}
\begin{enumerate}
  \setcounter{enumi}{1}
  \item $x^{(n+1)}=\argmin_{||x||_2=1} ||\cF x-y^{(n)}||^2$.
\end{enumerate}
\vspace{-2.5mm}
In practice, this modification might not always be possible, since the norm $\|x \|$
is not explicitly presented in $P_p$.
 However, in the classical setting of Fourier measurements and positive $x$, 
 this information is available.  Moreover, our empirical experience has been that using this regularization on well conditioned analysis dictionaries offers no benefit; in particular, it gives no benefit with random analysis matrices.

%add experiment showing doesn't help random?
\subsubsection{Phaselift and Phasecut}
Two recent algorithms \cite{phaselift} and \cite{1206.0102} are guaranteed 
with high probability to solve the (classical)  problem of recovering the phase 
of a complex signal from its modulus, given enough random measurements.  
In practice both perform better than the greedy alternating minimization.  
However, it is not obvious to us how to adapt these algorithms to the $\ell_p$ setting.

\subsubsection{Nearest neighbors regression}
\label{sec:nn_regress}
We would like to use the same basic algorithm for all settings to get an idea 
of the relative difficulty of the recovery problem for different $p$, but also would 
like an algorithm with good recovery performance. 
If our algorithm simply returns poor 
results in each case, differences between the cases might be masked.

The alternating minimization can be very effective when well initialized.  
When given a training set of the data to recover, we use a simple regression to find $x_0$.  
Fix a number of neighbors $q$ (in the experiments below we use $q=10$, and suppose $X$ is the training set).
Set $G=P_p(X)$, and for a new point $x$ to recover from $P_p(x)$, find the $q$ nearest neighbors in 
$G$ of $P_p(x)$, and take their principal component to serve as $x_0$ in the alternating minimization algorithm.  We use the fast neighbor searcher from \cite{vedaldi08vlfeat} to accelerate the search.

\subsection{Experiments}
We discuss results on the MNIST dataset, available at \url{http://yann.lecun.com/exdb/mnist/}, and on $16\times 16 $ patches drawn from the VOC dataset, available at \url{http://pascallin.ecs.soton.ac.uk/challenges/VOC/voc2012/}.  For each of these data sets, we run experiments with random dictionaries and adapted dictionaries.  
We also run experiments where the data and the dictionary are both Gaussian i.i.d.; in this case, we do not use adapted dictionaries. 

  The basic setup of the experiments in each case is the same: 
we vary the number of measurements (that is, number of pools) over some range, and attempt to recover the original signal from the $\ell_p$ pooled measurements,
using various methods.
 We record the average angle between the recovered signal $r$ and the original $x$, that is, we use $|r^Tx|^2/(||r||^2||x||^2)$ as the measure of success in recovery.  

In each case the random analysis dictionary $\cF$ is built by fixing a size parameter $m$, and generating a Gaussian i.i.d. matrix $\cF_0$ of size $2m\times n$, where $n=100$ for MNIST, and $n=256$ for VOC.  Each pair of rows of $\cF_0$ is then orthogonalized to obtain $\cF$; that is we use groups of size $2$, where the pair of elements in each group are orthogonal.  This allows us to use standard phase recovery software in the $\ell_2$ case to get a baseline.  We used the ADMM version of phaselift from \cite{conf/nips/OhlssonYDS12}  and the phasecut algorithm of \cite{1206.0102}. For all of our data sets, the latter gave better results (note that phasecut can explicitly use the fact that the solution to the problem is real, whereas that version of phaselift cannot), so we report only the phasecut results.   

In the experiments with adapted dictionaries, the dictionary is built using block OMP and batch updates with a K-SVD type update \cite{Aharon:2006:SAD:2197945.2201437};  in this case, $\cF$ is the transpose of the learned dictionary.  We again use groups of size $2$ in the adapted dictionary experiments.

We run two sets of experiments with Gaussian i.i.d.~data and dictionaries, with $n=20$ and $n=40$. 
 We consider $m$ in the range from $n/2$ to $8n$.  On this data set, phaselift outperforms alternating minimization; see the supplementary material.

%\subsubsection{MNIST}
%\label{mnist_experiments}
%The MNIST data set consists of 70000 $28\times 28$ images of handwritten digits.  The data set is organized into a standard training set of 60000 points and test set of 10000 points.  
For MNIST, we use the standard training set projected to $\R^{100}$ via PCA, and we let the number of dictionary elements range from 60 to 600 (that is, 30 to 300 measurements).   On this data set, alternating minimization with nearest neighbor initialization gives exact reconstruction by $130$ measurements; for comparison, Phaselift at $m=130$ has mean square angle of $.48$; see the supplementary material.

%\subsubsection{image patches}
%\label{image_patch_experiments}
We draw approximately 5 million $16\times 16$ grayscale image patches from the PASCAL VOC data set; these are sorted by variance, and the largest variance 1 million are kept.  The mean is removed from each patch.  These are split into a training set of 900000 patches and a test set of 100000 patches.  In this experiment,  we let $m$ range from 30 to 830.
 On this data set, by $m=330$ measurements, alternating minimization with nearest neighbor initialization recovers mean angle of $.97$; for comparison, Phaselift at $m=330$ has mean angle of $.39$; see the supplementary material.
\begin{figure*}[!ht]
\begin{center}
 \includegraphics[width=.3\linewidth,height=.15\linewidth]{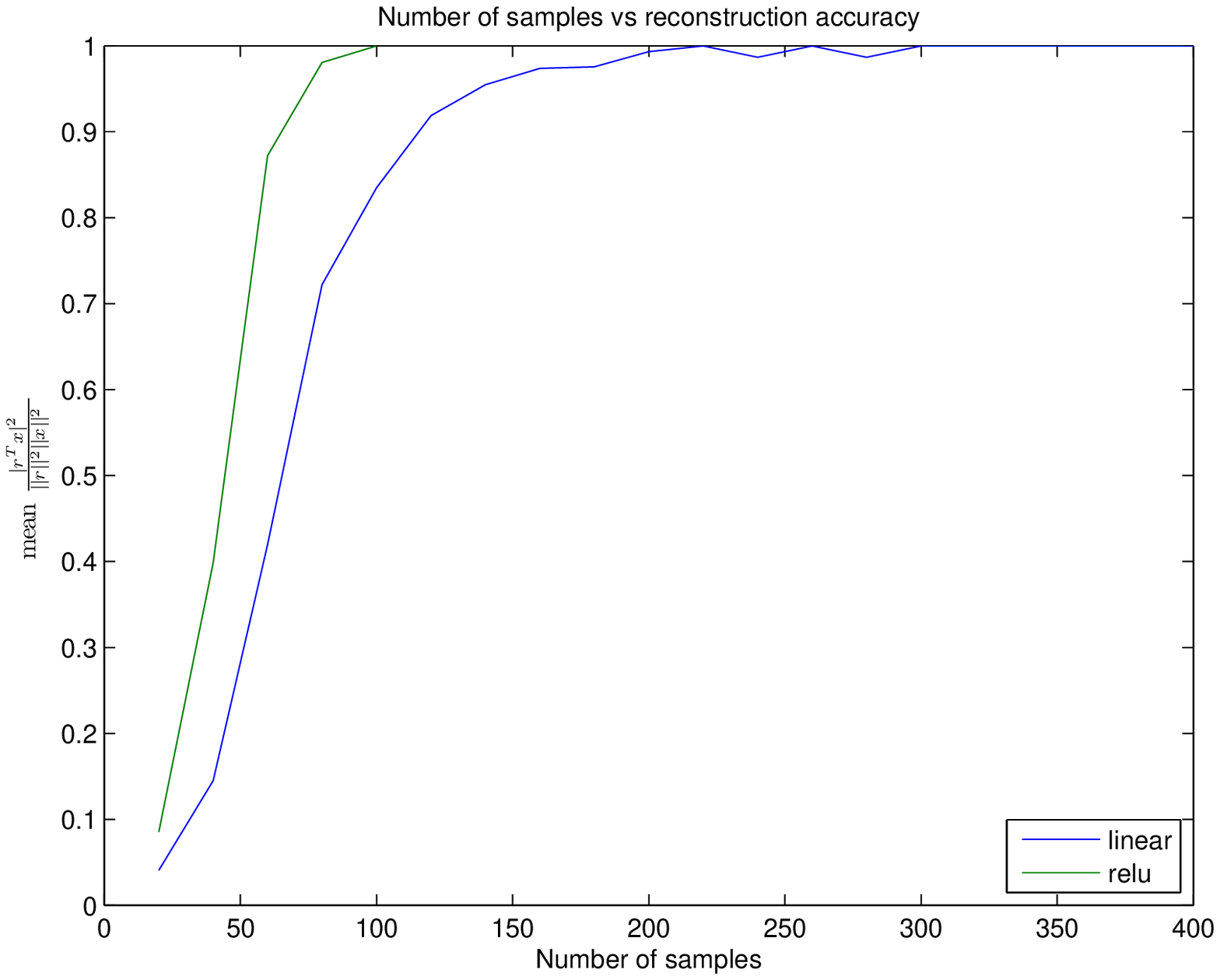} 
\includegraphics[width=.3\linewidth,height=.15\linewidth]{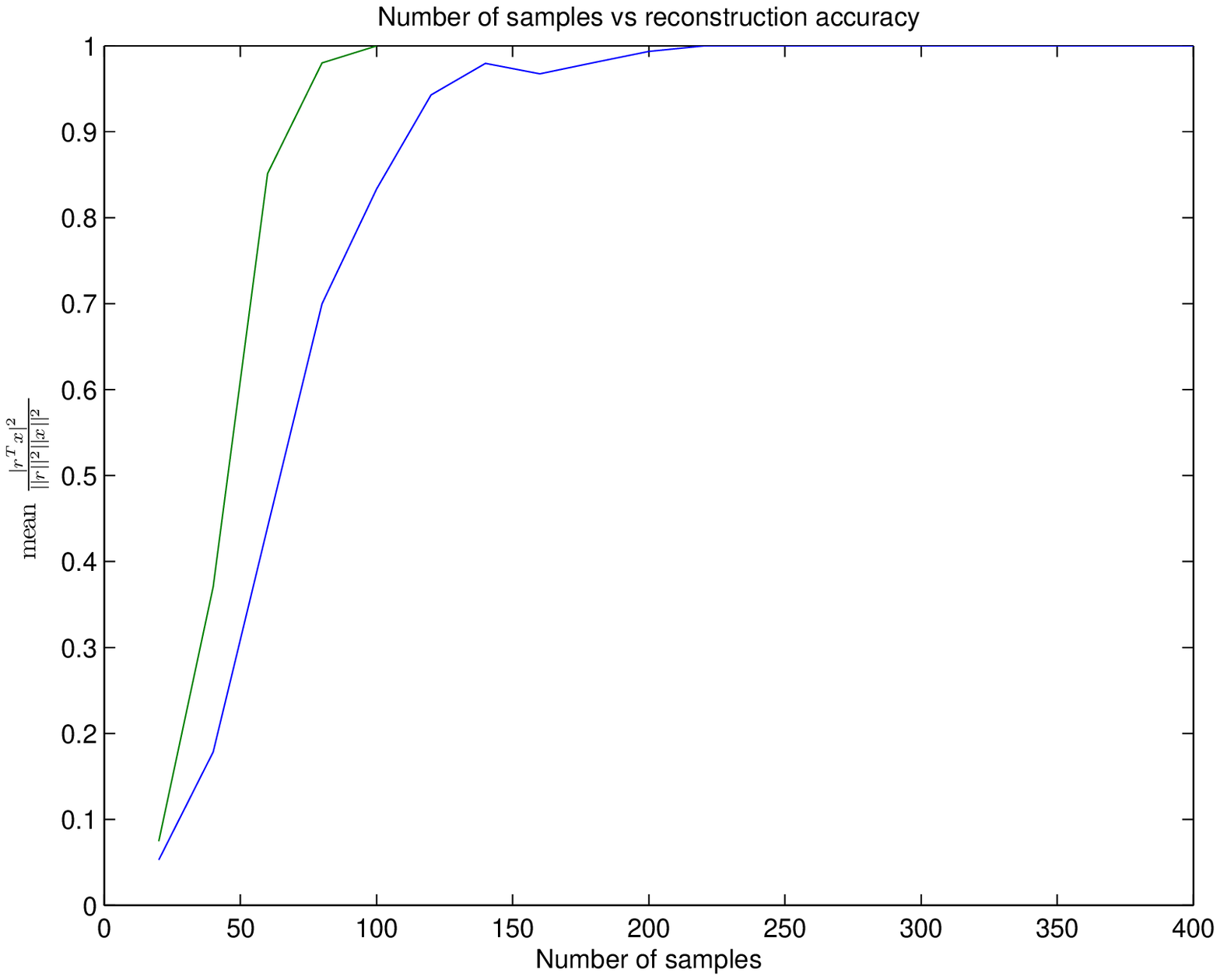}
 \includegraphics[width=.3\linewidth,height=.15\linewidth]{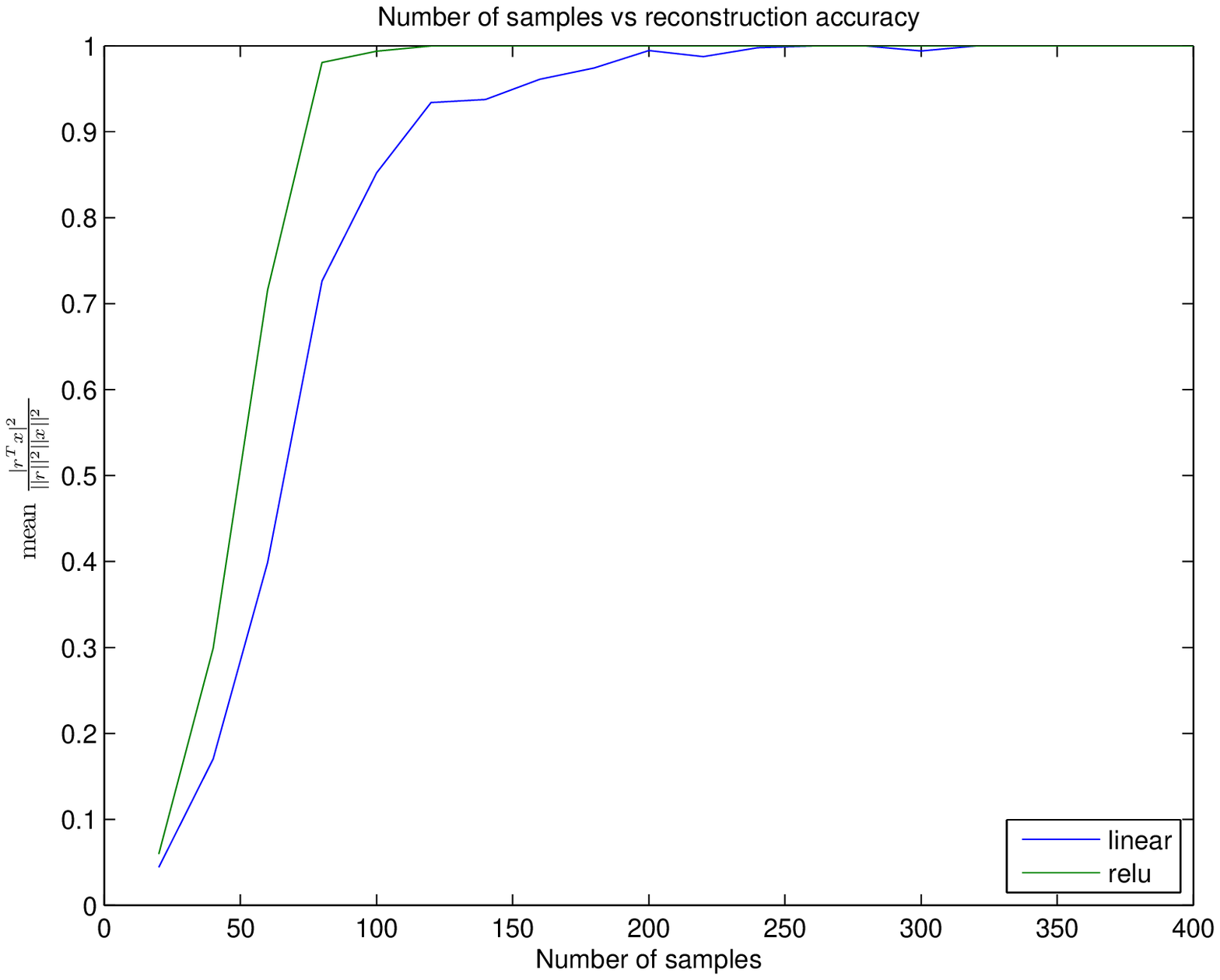}
\end{center}
\caption{Average recovery angle using alternating projections on random data; each $x$ is Gaussian i.i.d. in $\R^{40}$.   The vertical axis measures the average value of $|r^Tx|^2/(||r||^2||x||^2)$, where $r$ is the recovered vector, over 50 random test points.  The horizontal axis is the number of measurements (the size $m$ of the analysis dictionary is twice the $x$ axis in this experiment).  The leftmost figure is $\ell_1$ pooling, the middle $\ell_2$, and the right max pooling.  %In the top row each $x$ is Gaussian i.i.d. in $\R^{20}$, in the bottom row, in $\R^{40}$.  
The dark blue curve is alternating minimization, and the green curve is alternating minimization with half rectification; both with random initialization.\label{f:random_random}}
\end{figure*}
%%%%%%%%%%%%%%%%%%%%%%%%%%%%%
\begin{figure*}
\centering
\subfigure[MNIST, random filters]{%\begin{center}
\includegraphics[width=.3\linewidth,height=.2\linewidth]{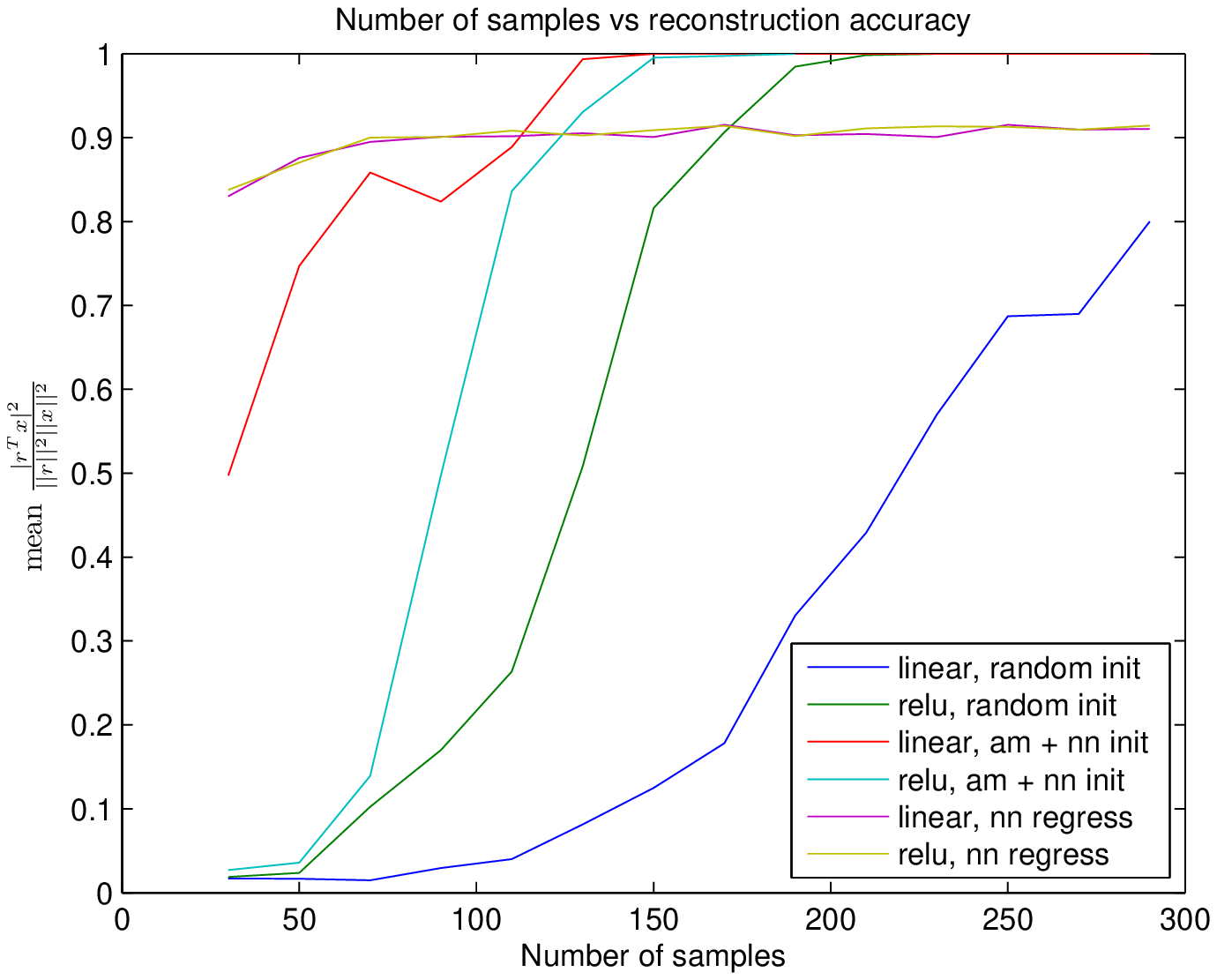} 
\includegraphics[width=.3\linewidth,height=.2\linewidth]{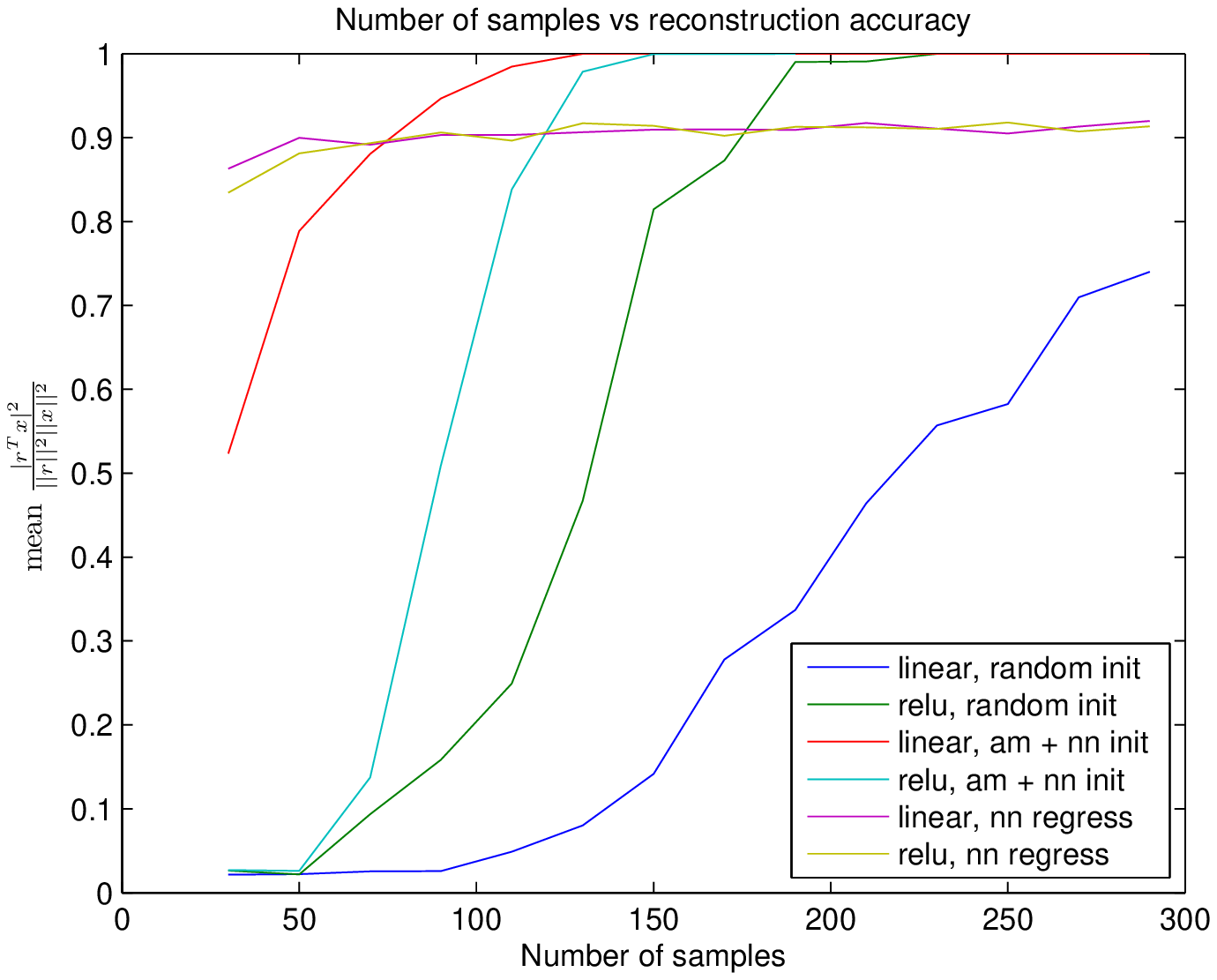}
 \includegraphics[width=.3\linewidth,height=.2\linewidth]{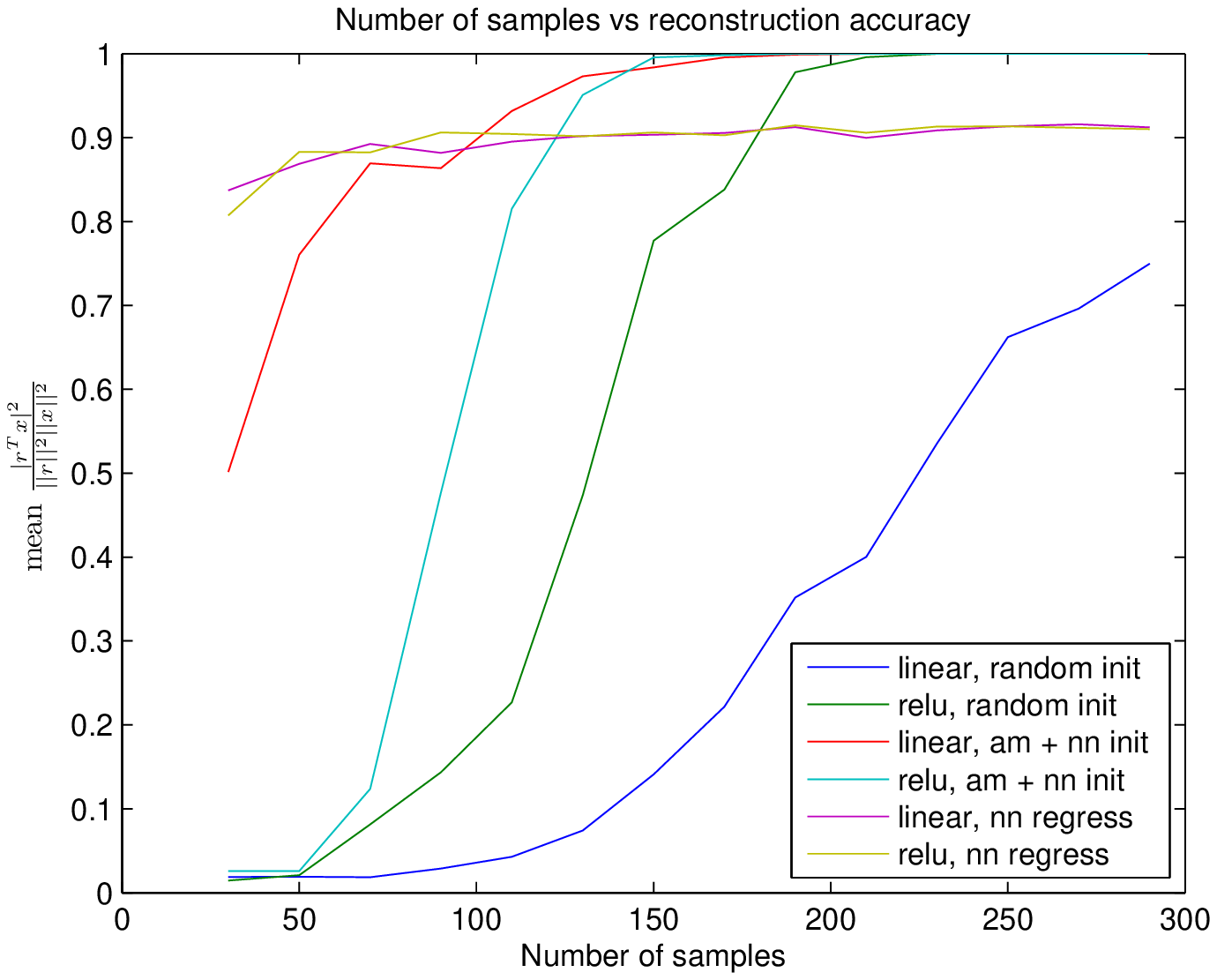}}
 \subfigure[MNIST, adapted filters]{
 \includegraphics[width=.3\linewidth,height=.2\linewidth]{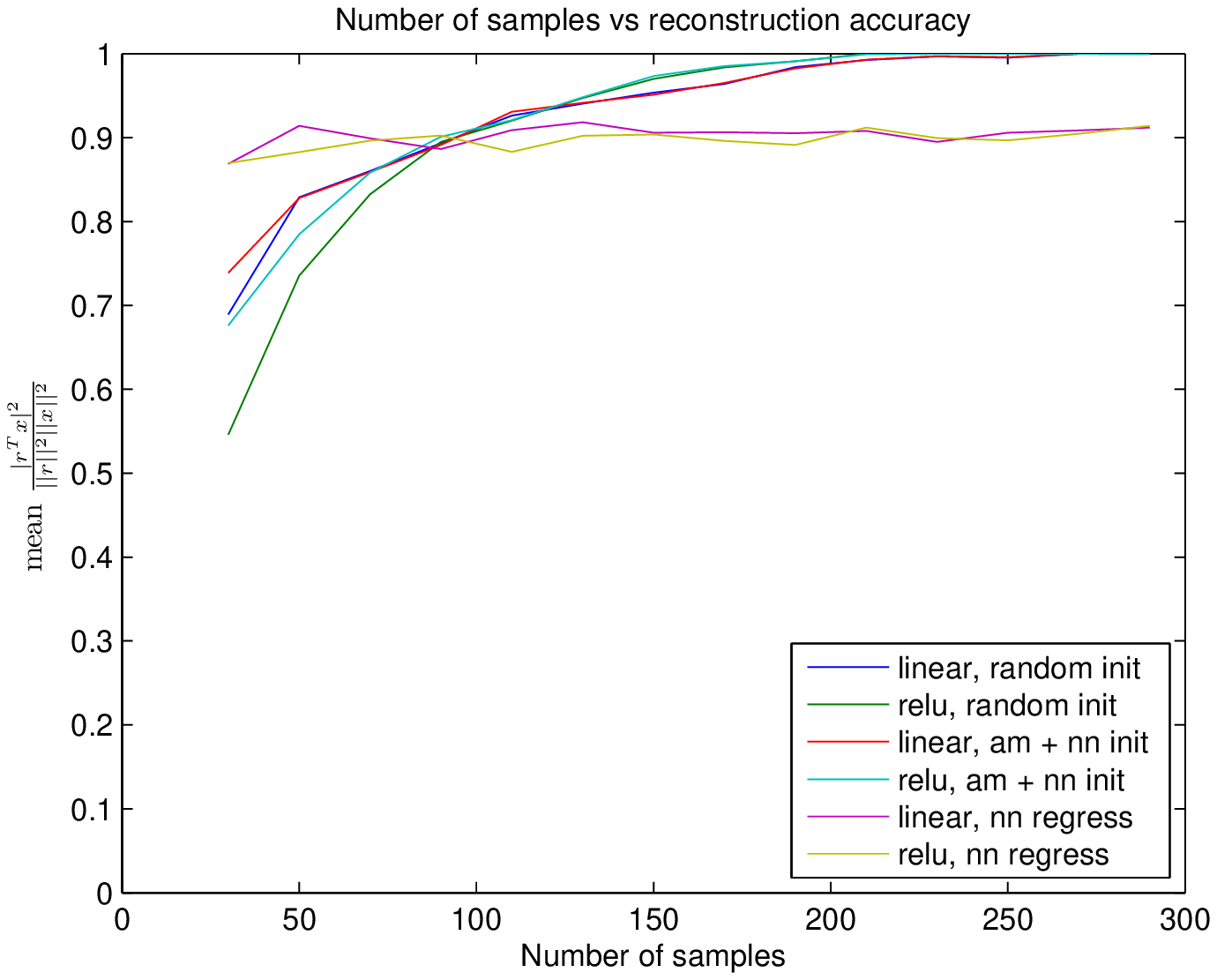} 
\includegraphics[width=.3\linewidth,height=.2\linewidth]{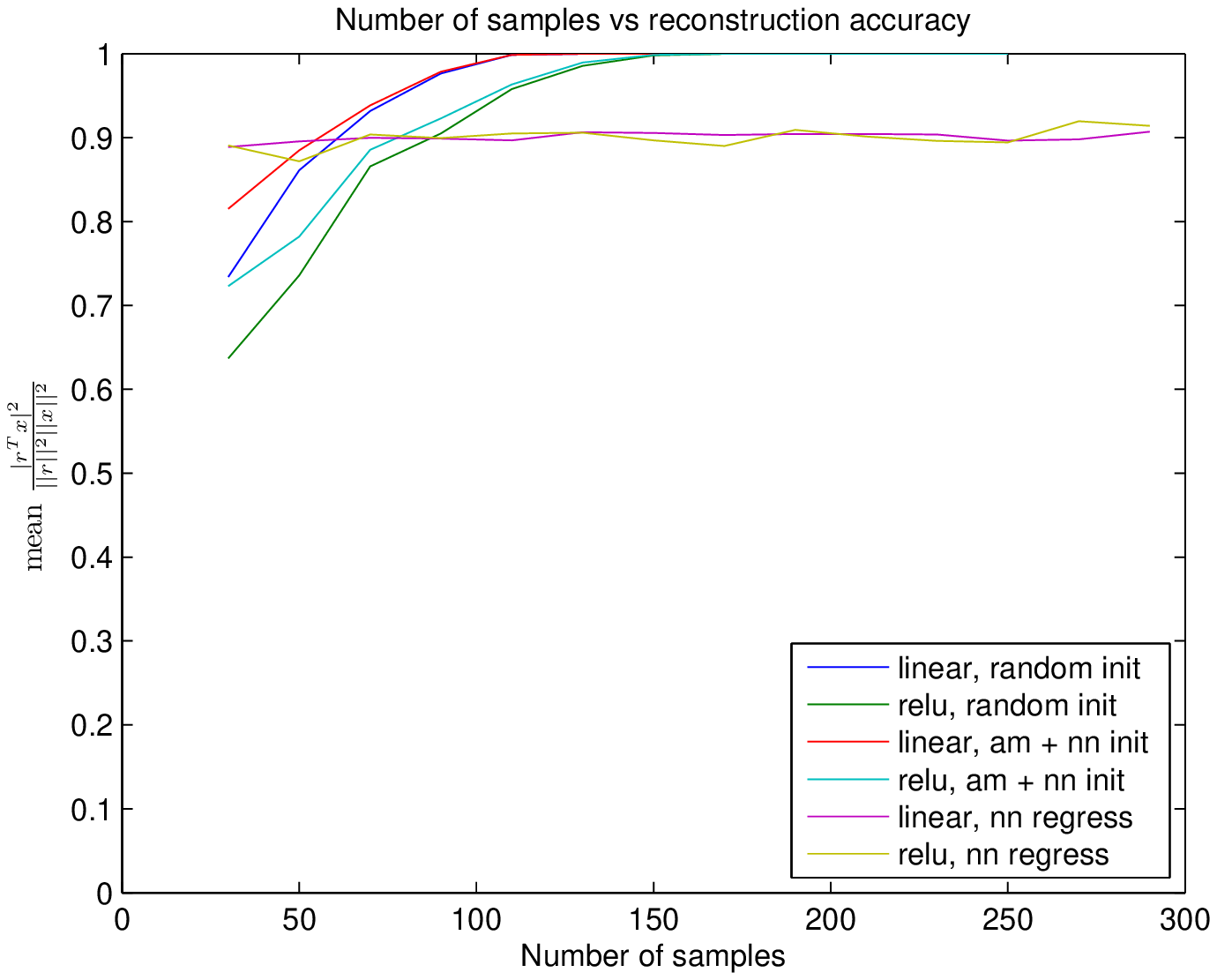}
 \includegraphics[width=.3\linewidth,height=.2\linewidth]{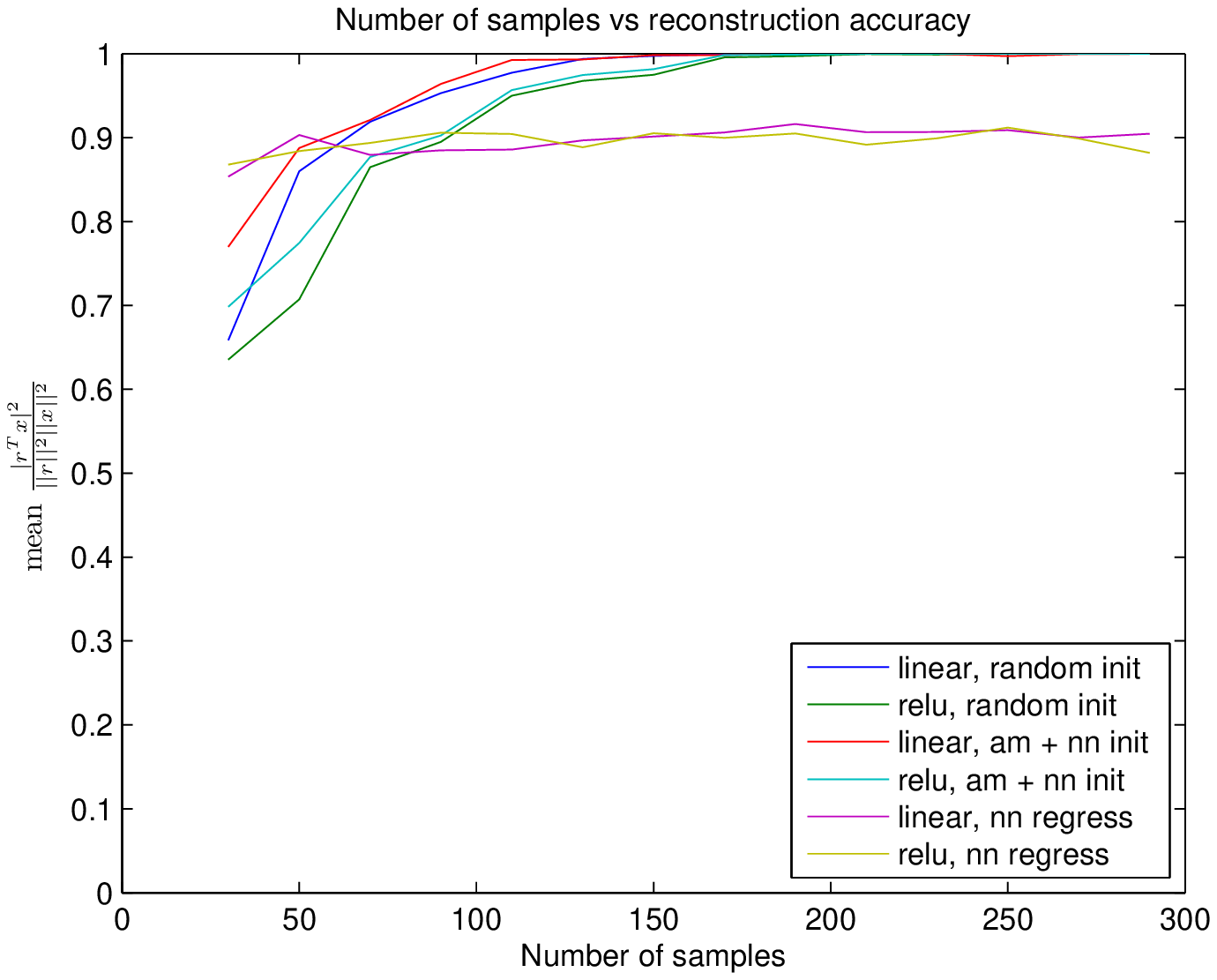}
%\end{center}
}
%\caption{Average recovery angle using alternating projections on MNIST data points.   The vertical axis measures the average value of $|r^Tx|^2/(||r||^2||x||^2)$, where $r$ is the recovered vector, over 50 random test points.  The horizontal axis is the number of measurements (the size of the analysis dictionary is twice the $x$ axis in this experiment).  The leftmost figure is $\ell_1$ pooling, the middle $\ell_2$, and the right max pooling.  In the top row the analysis dictionary is Gaussian i.i.d.;  in the bottom row, it is generated by block OMP/KSVD with $5$ nozero blocks of size 2.  The dark blue curve is alternating minimization, and the green curve is alternating minimization with half rectification; both with random initialization.  The magenta and yellow curves are the nearest neighbor regressor described in \ref{sec:nn_regress} without and with rectification ; and the red and aqua curves are alternating minimization initialized via neighbor regression, without and with rectification.\label{f:mnist}}
%\end{figure*}
%%%%%%%%%%%%%%%%%%%%%%%%%%%%%%%%%%%
%\begin{figure*}
\subfigure[Image patches, random filters]{%\begin{center}
\includegraphics[width=.3\linewidth,height=.2\linewidth]{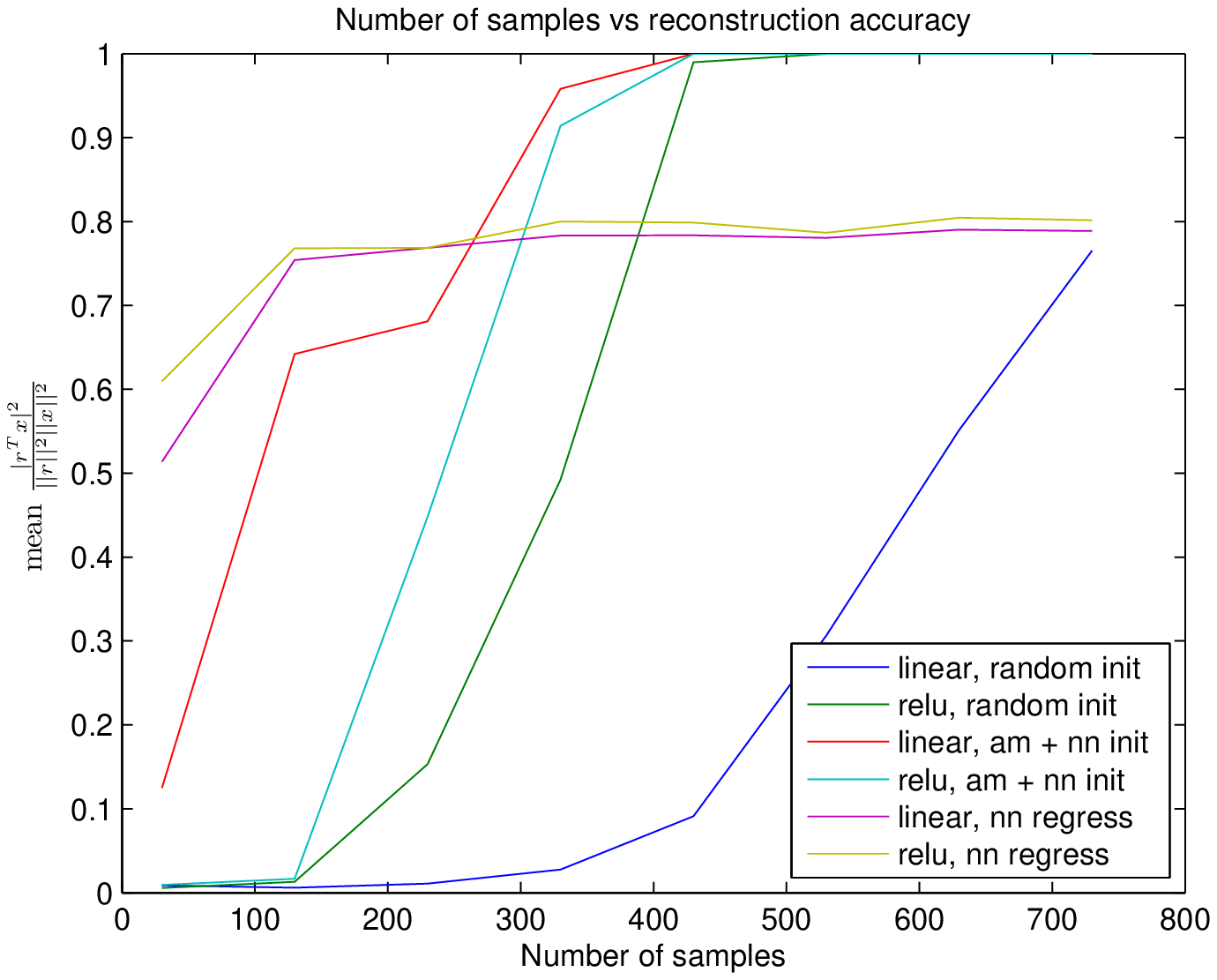} 
\includegraphics[width=.3\linewidth,height=.2\linewidth]{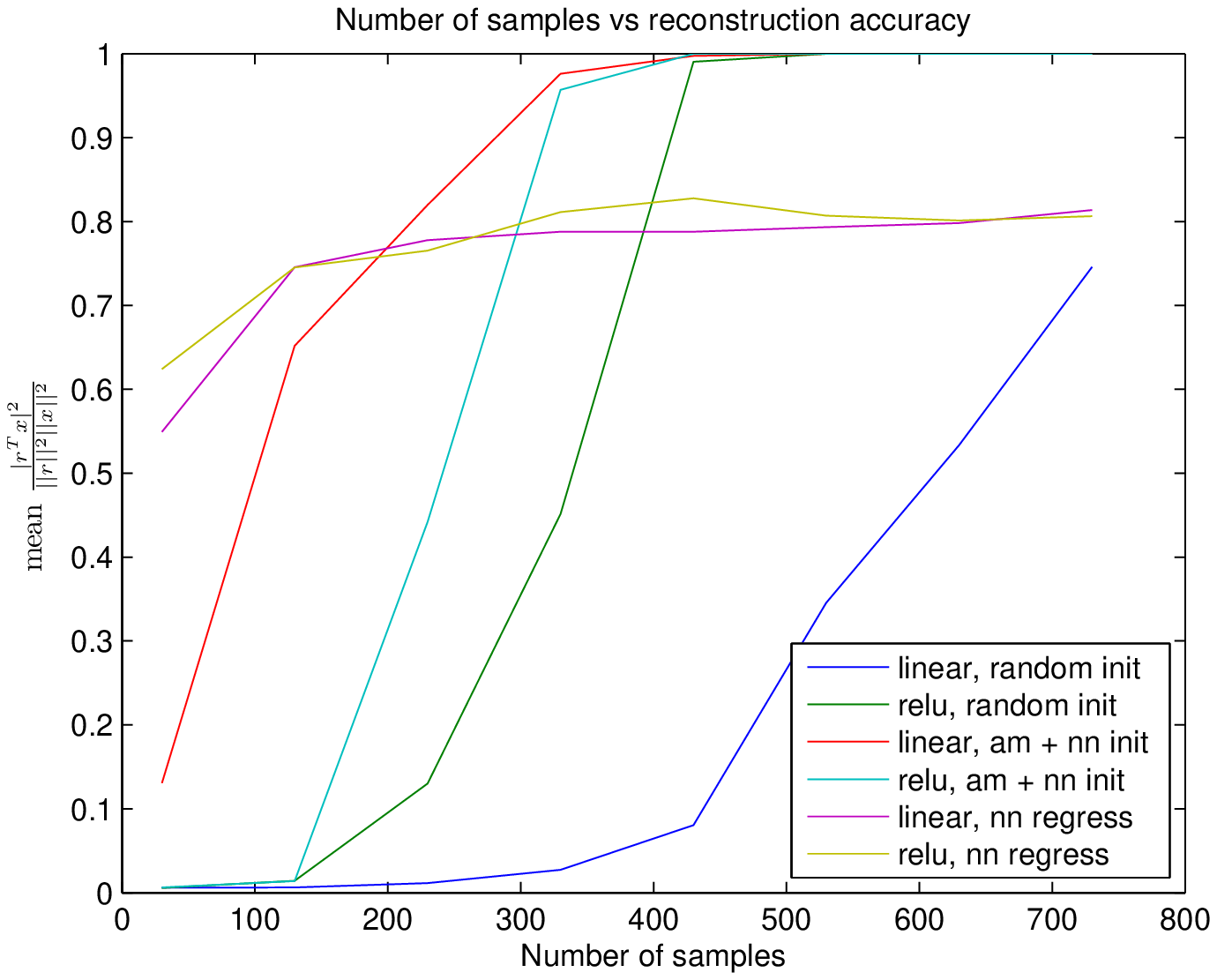}
 \includegraphics[width=.3\linewidth,height=.2\linewidth]{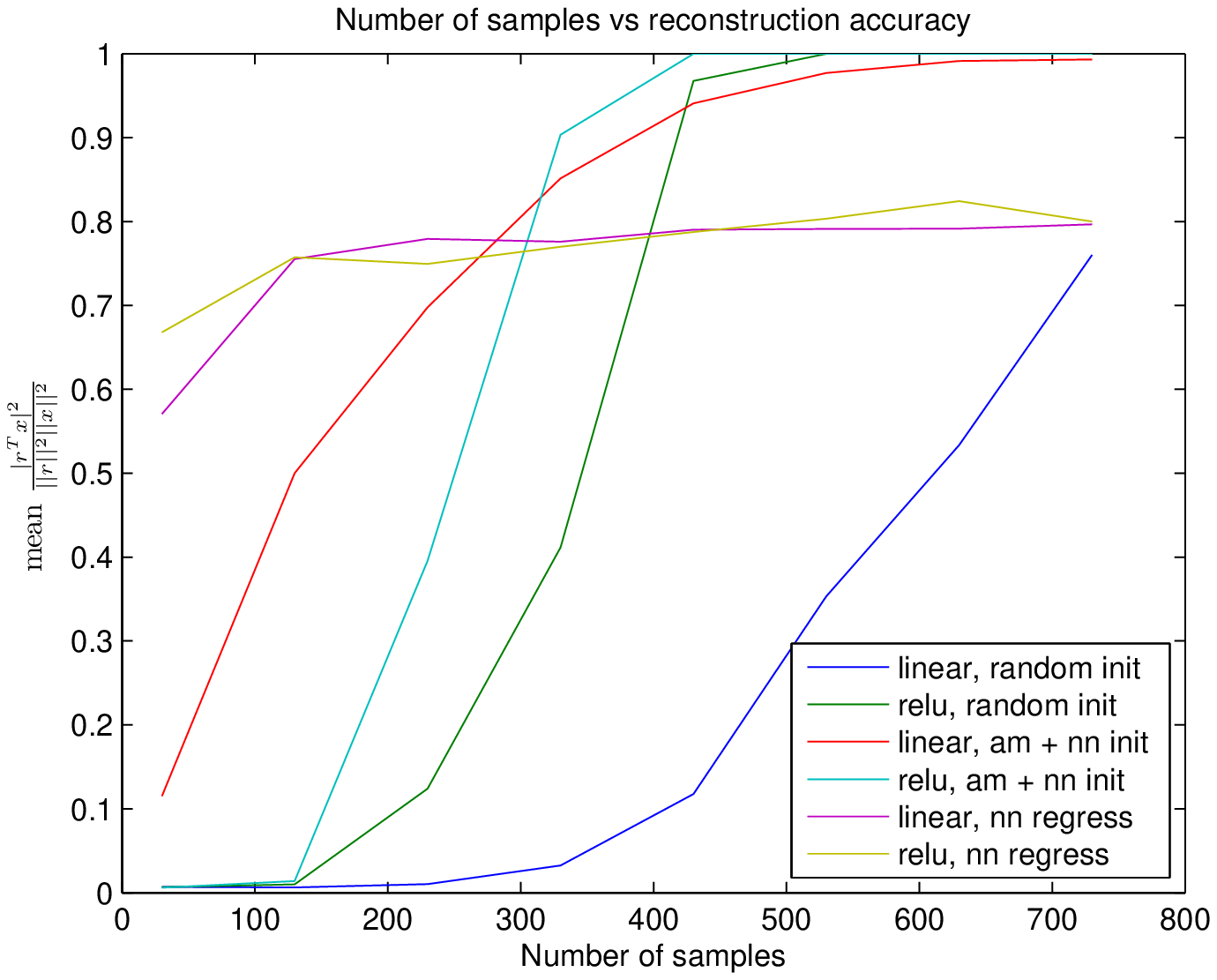}}
\subfigure[Image patches, adapted filters]{%
\includegraphics[width=.3\linewidth,height=.2\linewidth]{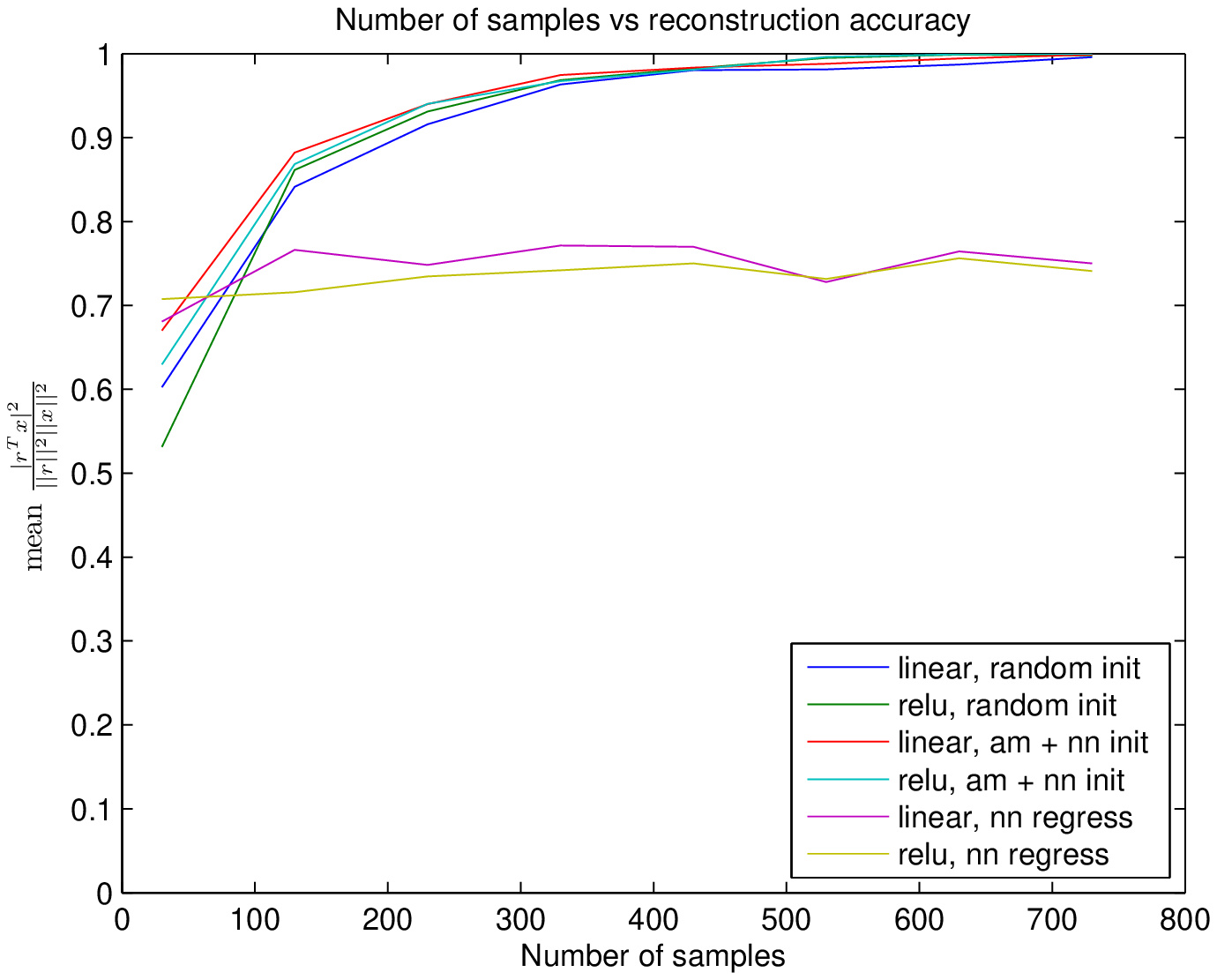} 
\includegraphics[width=.3\linewidth,height=.2\linewidth]{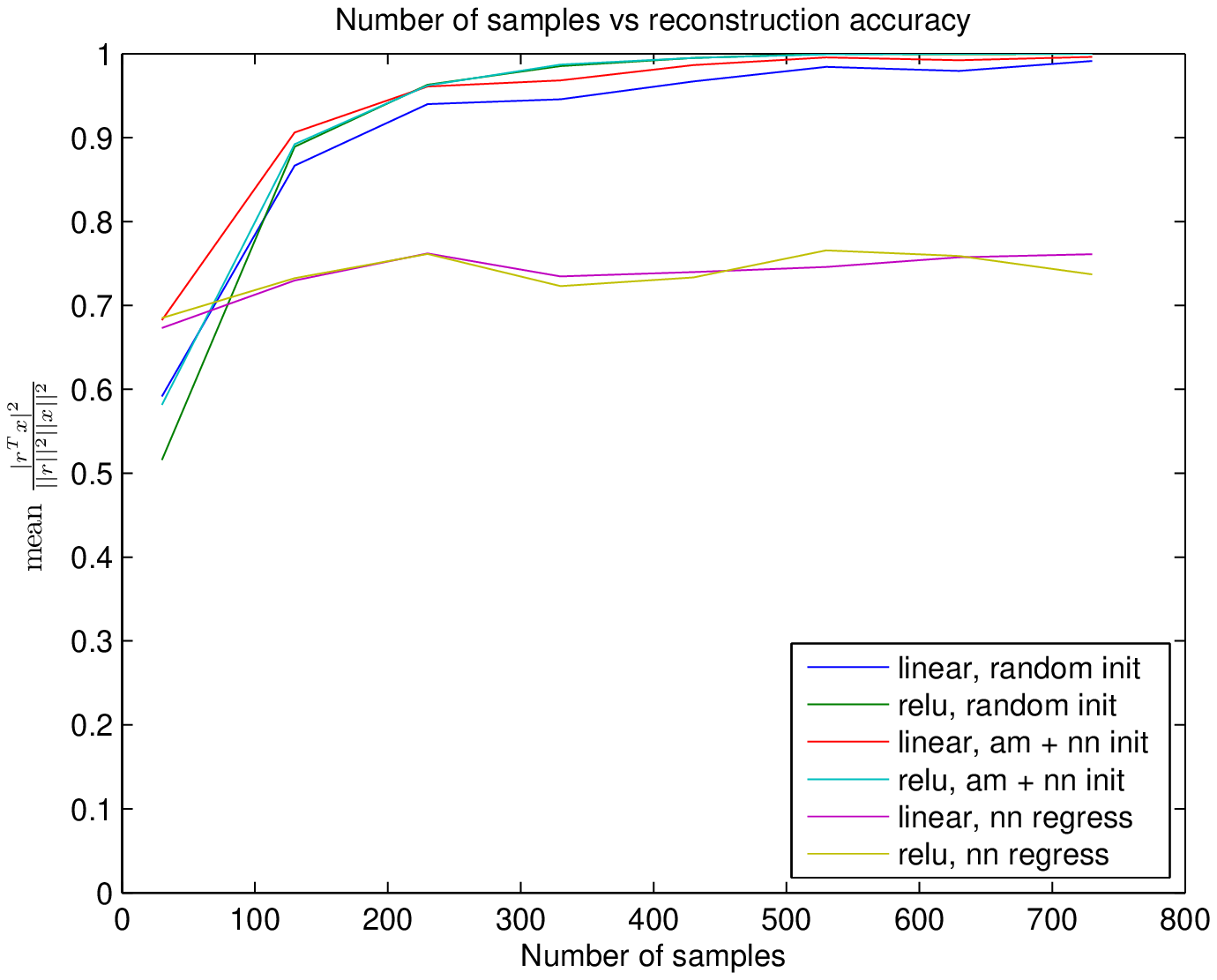}
 \includegraphics[width=.3\linewidth,height=.2\linewidth]{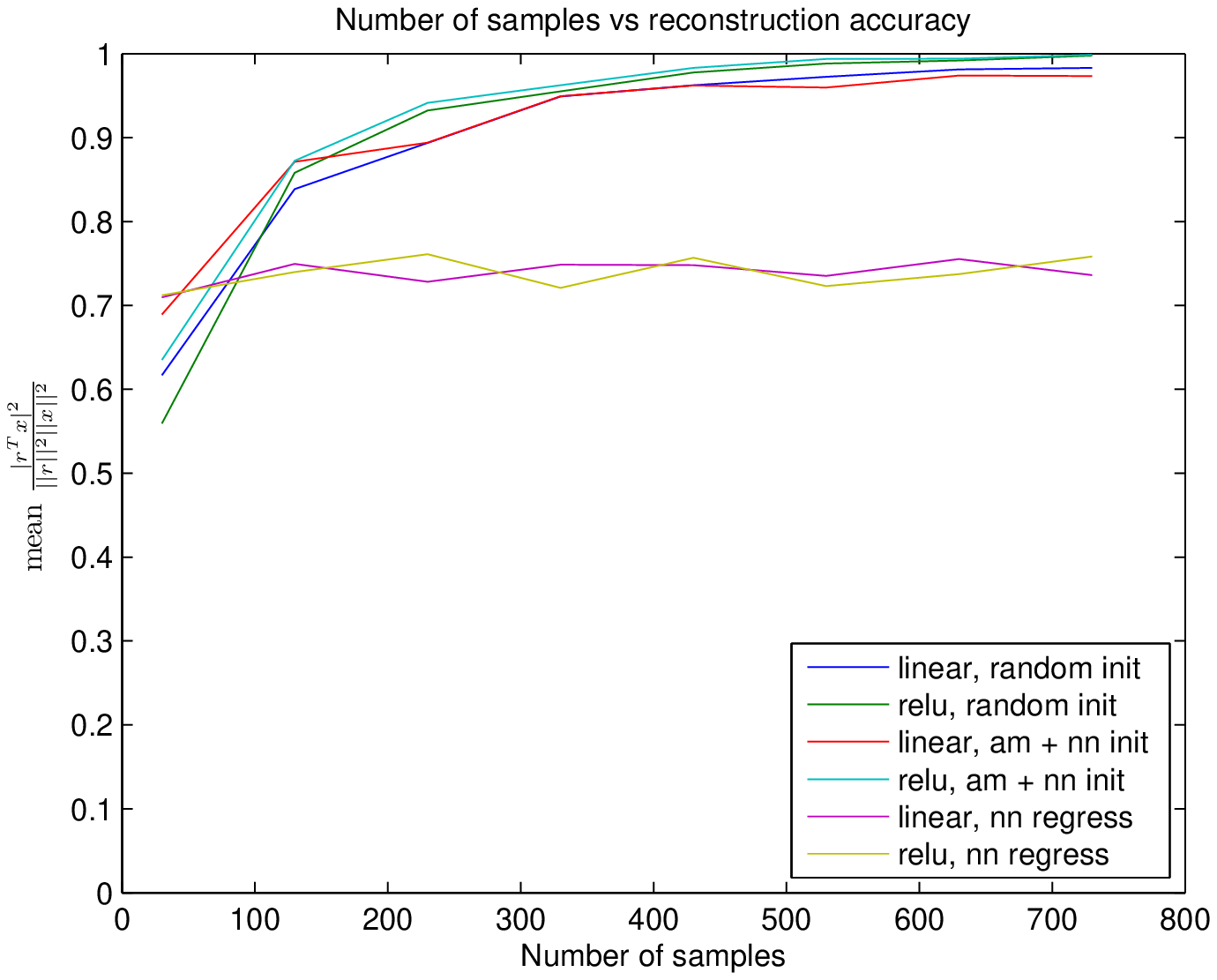}
 %\end{center}
 }
\caption{Average recovery angle using alternating projections on image patch data points.   The vertical axis measures the average value of $|r^Tx|^2/(||r||^2||x||^2)$, where $r$ is the recovered vector, over 50 random test points.  The horizontal axis is the number of measurements (the size of the analysis dictionary is twice the $x$ axis in this experiment).  The leftmost figure is $\ell_1$ pooling, the middle $\ell_2$, and the right max pooling.  In the top row of each pair of rows the analysis dictionary is Gaussian i.i.d.;  in the bottom row of each pair of rows, it is generated by block OMP/KSVD with $5$ nozero blocks of size 2. The dark blue curve is alternating minimization, and the green curve is alternating minimization with half rectification; both with random initialization.  The magenta and yellow curves are the nearest neighbor regressor  described in \ref{sec:nn_regress} without and with rectification; and the red and aqua curves are alternating minimization initialized via neighbor regression, without and with rectification.  See Section \ref{sec:experimental_analysis} for a discussion of the figures.\label{f:mnist_patches}}
\end{figure*}

\subsection{Analysis}
\label{sec:experimental_analysis}
The experiments show (see figures \ref{f:random_random} and \ref{f:mnist_patches}) that: 
\vspace{-3.5mm}
\begin{itemize}
\item For every data set, with random initializations and dictionaries, recovery  is easier with half rectification before pooling than without (green vs dark blue in figures).
\item $\ell_{\infty}$,  $\ell_{1}$,  and $\ell_{2}$ pooling appear roughly the same difficulty to invert, regardless of algorithm (each column of figures, corresponding to an $\ell_p$, is essentially the same). 
\item Good initialization improves performance; indeed, alternating minimization with nearest neighbor regression outperforms phaselift and phasecut (which of course do not have the luxury of samples from the prior, as the regressor does).  We believe this of independent interest.
\item Adapted analysis ``frames'' (with regularization) are easier to invert than random analysis frames, with or without regularization (the bottom row of each pair of graphs vs the top row of each pair in Figure \ref{f:mnist_patches}).
\end{itemize}
\vspace{-2.5mm}
Each of these conclusions is unfortunately only true up to the optimization method- it may be true that a different optimizer will lead to different results.
With learned initializations and alternating minimization, recovery results can be better without half rectification.  Note this is only up until the point where the alternating minimization gets better than the learned initialization without any refinement, and is especially true for random dictionaries.   The simple interpretation is that the reconstruction step 2 of the alternating minimization just does not have a large enough span with roughly half the entries removed; that is, this is an effect of the optimization, not of the difference between the difficulty of the problems. 

\section{Conclusion}
We have studied conditions under which neural network layers of the form \eqref{lppool} and \eqref{lprect} preserve signal
information. As one could expect, recovery
from pooling measurements is only guaranteed
 under large enough redundancy and configurations
 of the subspaces, which depend upon which $\ell_p$
 is considered.   We have proved conditions which bound the lower Lipschitz constants for these layers, giving quantitative descriptions of how much information they preserve.  Furthermore, we have given conditions under which modules with random filters are invertible.   We have also given experimental evidence that for both random and adapted modules, it is roughly as easy to invert $\ell_p$ pooling  with $p=1$, $2$, and $\infty$; and shown that when given training data, alternating minimization gives state of the art phase recovery with a regressed initialization.

However, we are not anywhere near where we would like to be  in understanding these systems, or even the invertibility of the layers of these systems.  This work gives little direct help  to a practicioner asking the question ``how should I design my network?''.  In particular, our results just barely touch on the distribution of the data; but the experiments make it clear 
(see also \cite{conf/nips/OhlssonYDS12}) that knowing more information about the data changes the invertibility of the mappings.   Moreover, preservation of information needs to be balanced against invariance, and the tension between these is not discussed in this work.   Even in the setting of this work, without consideration of the data distribution or tension with invariance, Proposition \ref{p2lips} although tight, is not easy to use, and even though we are able to use \ref{mpooling_prop} to get an invertibility result, it is probably not tight.

This work also shows there is much research to do in the field of algorithmic phase recovery. 
What are correct algorithms for $\ell_p$ inversion, perhaps with half rectification?   How can we best use knowledge of the data distribution for phase recovery, even for the well studied $\ell_2$ case?  Is it possible to guarantee that a well initialized alternating minimization converges to the correct solution?

%\clearpage

\bibliographystyle{plainnat}
\bibliography{refs}

\clearpage
\appendix
\section{Comparison between phaselift and the various alternating minimization algorithms}
Here we give a brief comparison between the phaselift algorithm and the algorithms we use in the main text.    Our main goal is to show that the similarities between the $\ell_1$, $\ell_2$, $\ell_{\infty}$ recovery results are not just due to the alternating minimization algorithm performing poorly on all three tasks;  however we feel that the quality of the recovery with a regressed initialization is interesting in itself, especially considering that it is much faster than either phaselift or phasecut.  

\begin{figure}[h!]
\includegraphics[width=.95\linewidth]{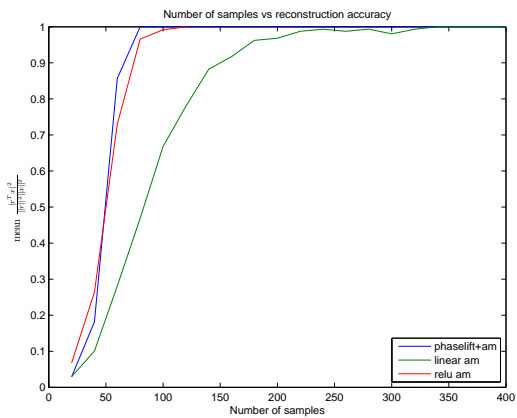} 
\caption{Average recovery angle using phaselift and alternating minimization on random data, Gaussian i.i.d. points in $\R^{40}$.  The blue curve is phaselift followed by alternating minimization; the green curve is alternating minimization, and the red is alternating minimization on pooling following half rectification.\label{fig:pcutvs_random}}
\end{figure}

\begin{figure}
\begin{center}
\includegraphics[width=.95\linewidth]{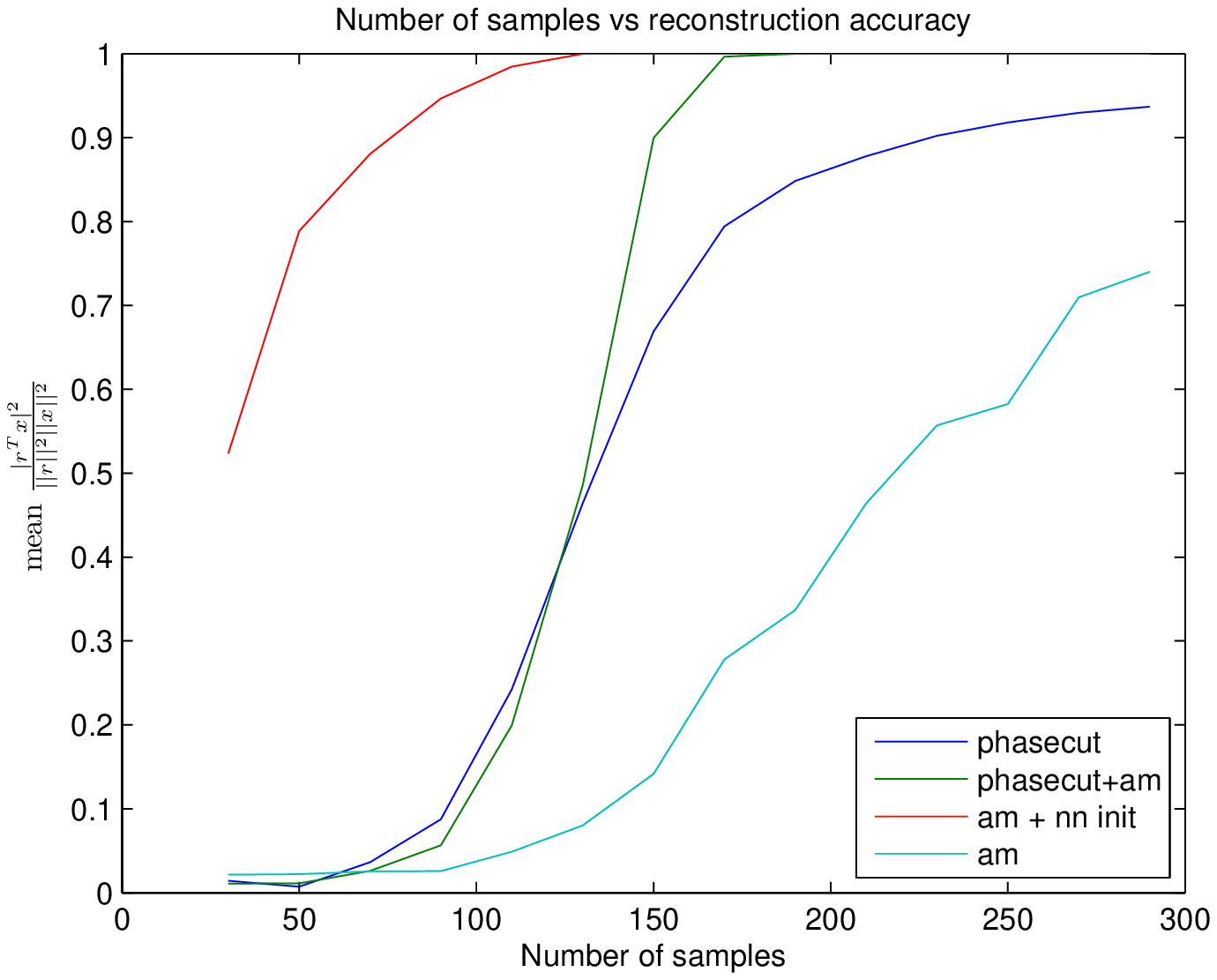}
\includegraphics[width=.95\linewidth]{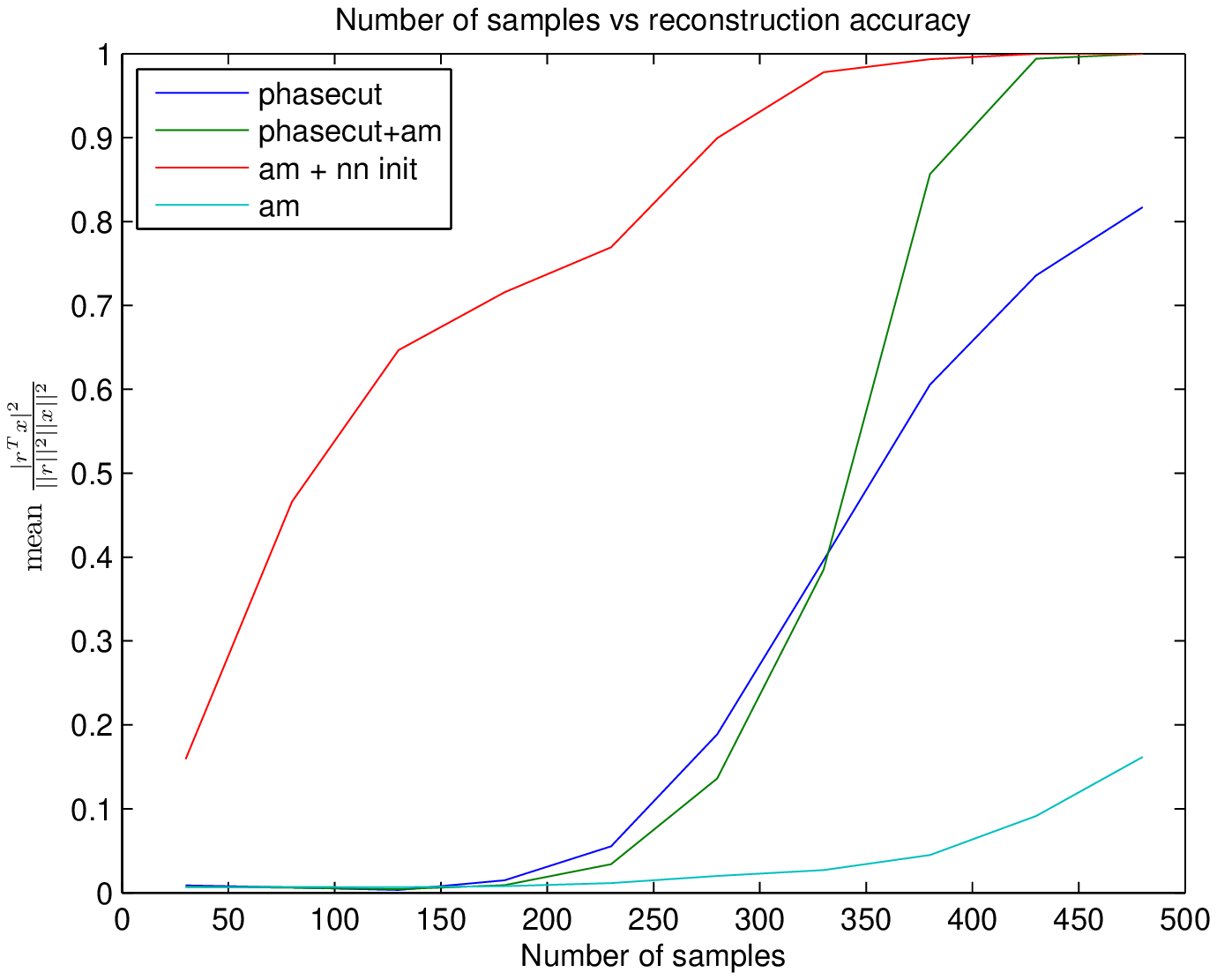}
 \end{center}
\caption{Average recovery angle using phaselift and alternating minimization on MNIST and patches data sets.    Top: MNIST digits, projected via PCA to $\R^{100}$.  Bottom: 16x16 image patches with mean removed.  The red curve is alternating minimization with nearest neighbor initialization, the green is alternating minimization initialized by phasecut (this is the recommended usage of phasecut), the blue is phasecut with no alternating minimization, and the aqua is alternating minimization with a random initialization. \label{fig:pcutvs_data}}
\end{figure}

In figures \ref{fig:pcutvs_random}, and \ref{fig:pcutvs_data} we compare phaselift against alternating minimization with a random initialization and alternating minimization with a nearest neghbor/locally linear regressed initialization.  Because we are comparing against phasecut, here we only show inversion of $\ell_2$ pooling.

In figure of \ref{fig:pcutvs_random}, we use random data and a random dictionary.  As the data has no structure, we only compare against random initialization, with and without half rectification.  We can see from figure \ref{fig:pcutvs_random} in this case, where we do not know a good way to initialize the alternating minimization, alternating minimization is significantly worse than phasecut.   On the other hand, recovery after rectified pooling with alternating minimization does almost as well as phasecut. 

 In the examples where we have training data, shown in figure \ref{fig:pcutvs_data}, alternating minimization with the nearest neighbor regressor (red curve) performs significantly better than phasecut (green and blue curves).  Of course phasecut does not get the knowledge of the data distribution used to generate the regressor.  

\clearpage
 \section{Proofs of results in Section \ref{sect2}}
\label{proofs}
\subsection{Proof of Proposition \ref{halfrect_inj}}

Let us first show that $A_0 > 0$ is sufficient 
to construct an inverse of $M_\alpha$.
Let $x \in \R^N$. By definition, 
the coordinates of $M_\alpha(x) > \alpha$  correspond to 
$$s(x) = \{ i \,\,s.t.\, \langle x,f_i \rangle > \alpha_i\} \subset \{1,\dots,M\}~,$$
which in particular implies that $x$ is known to lie in $V_{S(x)}$, 
the subspace generated by $s(x)$.
But the restriction $\cF_{s(x)}$ is a linear operator, which can be inverted in
$V_S$
as long as $\lambda_{-} (\restr{\cF_{s(x)}}{V_S}) \geq A_0 > 0$.

Let us now show that $A_0 > 0$ is also 
necessary. Let us suppose that for some $S$,
 $\cF_{S}$ is such that $\lambda_{-}(\restr{\cF_{S}}{V_S})=0$. 
%Let $\gamma(x)=\min_i | \langle x , f_i \rangle |$. 
%If $\gamma(e)=0$, we replace $e$ by 
%a perturbation $e'$ such that $S_e=S_{e'}$
%and $\gamma(e')>0$. If $W$ has 
%rank $N$, such perturbations
%always exist; if $\mbox{rank}(W) <N $ 
%then $R_W$ is not invertible and we are done.
%Since $W_{S_e}$ does not satisfy (\ref{cond_inv}), 
It results that there exists $\eta \in V_S$ such
that $\|\eta\| > 0$ but $ \| \cF_{S} \eta \|=0$.
Since $S$ is a cone, we can find $x \in S$ and
$\epsilon \neq 0$ small enough
such that $x+\epsilon e \in S$.
It results that $M_\alpha(x) =M_\alpha(x+\epsilon e)$
which implies that $M_\alpha$ cannot be injective.

Finally, let us prove (\ref{halfrect_lips}). 
If $x,x'$ are such that $S=s(x)=s({x'})$, then 
$$\| M_\alpha(x) - M_\alpha(x') \| = \| \cF_{S} (x-x') \| \geq A_0 (x-x')~.$$
If $s(x) \neq s(x')$, we have that
$|M_\alpha(x)_i - M_\alpha(x')_i |= |\langle x-x', f_i \rangle |$ if $i\in s(x) \cap s({x'})$ 
and $|M_\alpha(x)_i - M_\alpha(x')_i | \geq | \langle x-x', f_i \rangle |$ 
if $i\in s(x) \cup s({x'})$, $i \notin s(x) \cap s({x'})$.  
It results that 
$$\| M_\alpha(x) - M_\alpha(x') \| \geq \| \cF_{s(x) \cup s({x'})}(x-x') \| \geq A_0 \|x-x'\|~.$$

 $\square$. 
 
%\subsection{Proof of Corollary \ref{rectif_corollary}}

\subsection{Proof of Proposition \ref{p2lips}}

The upper Lipschitz bound is obtained by observing
that, in dimension $d$, 
$$\forall\,y \in \R^d~,~\|y \|_1 \leq \sqrt{d} \| y \|_2~,~\| y \|_\infty \leq d \|y \|_2~.$$
It results that 
\begin{eqnarray}
\label{pizza3}
\| P_p(x) - P_p(x') \| &\leq& \alpha_p \| P_2(x) - P_2(x') \|  \\
&=& \alpha_p \| M(x)-M(x') \| \leq \alpha_p \lambda_{+}(\cF)~. \nonumber
\end{eqnarray}
Let us now concentrate on the lower Lipschitz bound.
Given $x,x' \in \R^n$, we first 
consider a rotation $\tilde{\cF}_k $ on each subspace $\cF_k$ such 
that $\langle x, \tilde{f}_{k,j} \rangle = \langle x', \tilde{f}_{k,j} \rangle =0 $ 
for $j>2$, which always exists. If now we modify $\tilde{\cF}_k$
 by applying a rotation of the remaining two-dimensional subspace such 
that $x$ and $x'$ are bisected, one can verify that 
\begin{eqnarray*}
\left(\| \cF_k x \|_2 - \| \cF_k x' \|_2\right )^2 &=& (\| \widetilde{\cF}_k x \|_2 - \| \widetilde{\cF}_k x' \|_2)^2 \\
&=& (|\langle x , \tilde{f}_{k,1} \rangle | - |\langle x' , \tilde{f}_{k,1} \rangle | )^2 \\
&& +  (|\langle x , \tilde{f}_{k,2} \rangle | - |\langle x' , \tilde{f}_{k,2} \rangle | )^2~,
\end{eqnarray*}
which implies, by denoting $M(x)=( |\langle x, \widetilde{f}_{k,j} \rangle| )_{k,j}$,
 that $\| P_2(x) - P_2(x') \| = \| M(x) - M(x') \|$. Since $\widetilde{\cF} \in \cQ_2$, 
 it results from Proposition \ref{balanprop} that 
\begin{eqnarray}
\label{pizza4}
 \| P_2(x) - P_2(x') \| &\geq & d(x,x') \min_{S \subset \{1\dots m\}} \sqrt{ \lambda_{-}^2( \widetilde{\cF}_S) + \lambda_{-}^2 (\widetilde{\cF}_{{S}^c})} \nonumber \\ 
&\geq & d(x,x') A_2 ~~\square.
\end{eqnarray}

 \subsection{Proof of Corollary \ref{rectpoolcor}}

Given $x,x'$, let $I$ denote the groups $I_k$, $k\leq K$ 
such that $S_x \cap S_{x'} \cap I_k = I_k$. 
It results that 
{
\[\| R_p (x) - R_p(x') \|^2\]\[=\sum_{k \in I} |R_p(x)_k - R_p(x')_k|^2 + \sum_{k \notin I} |R_p(x)_k - R_p(x')_k|^2\]
\[\geq   \sum_{k \in I} |R_p(x)_k - R_p(x')_k|^2 + \sum_{k \notin I} (\|\restr{M_0(x)}{I_k} - \restr{M_0(x')}{I_k} \|)^2.\]

%\begin{eqnarray*}
%%\label{banana1}
%\| R_p (x) - R_p(x') \|^2 &=& \sum_{k \in I} |R_p(x)_k - R_p(x')_k|^2 + \sum_{k \notin I} |R_p(x)_k - R_p(x')_k|^2 \nonumber \\
%&\geq&  \sum_{k \in I} |R_p(x)_k - R_p(x')_k|^2 + \sum_{k \notin I} (\|\restr{M_0(x)}{I_k} - \restr{M_0(x')}{I_k} \|)^2 ~.
%\end{eqnarray*}
}
On the groups in $I$ we can apply the same arguments as in 
theorem \ref{p2lips}, and hence find a frame $\tilde{\cF}$ 
from the family $\widetilde{\cQ}_{p,x,x'}$ such that 
$$\|R_p(x) - R_p(x') \|_I = \| M(x) - M(x') \|~,$$
with $M(x)=( |\langle x, \widetilde{f}_{k,j} \rangle| )_{k\in I,j}$ 
and $\{\widetilde{f}_{k,j} \} \in \widetilde{\cQ}_{p,x,x'}$.
Then, by following the same arguments used previously,
it results from the definition of $\tilde{A}_p$ %and (\ref{banana1})
 that
$$\| R_p (x) - R_p(x') \| \geq \tilde{A}_p d(x,x')~.$$

Finally, the upper Lipschitz bound is obtained by noting that
$$\| M_\alpha(x) - M_\alpha(x') \| \leq \| \cF(x-x') \|~,$$
and using the same argument as in (\ref{pizza3}) $\square$.

% Finally, let us verify that the rectified pooling operators $R_p$ 
% satisfy the same Lipschitz bounds as their counterparts $P_p$.
%For each pool $k \leq K$, 
% the thresholding on the coefficients $y_j= \langle x, f_{k,j} \rangle $ restricts  
%the $\ell_p$ balls $\{ y \,; \|y\|_p=r\} $ into the sets $\{ y\,; \,\|y \|_p=r\, y_j \geq \alpha_j \}$.
%Since in the proof of (\ref{p2_lips_conds}) 
%the changes of basis were performed independently for each pool, 
%it follows that one can also find frames $\widetilde{\cF}_k$ for each $k$
%such that $ \| R_p(x) - R_p(x') \| = \| M(x) - M(x') \|$, 
%with $M(x)=( |\langle x, \widetilde{f}_{k,j} \rangle| )_{k,j}$ 
%

%

%
\subsection{Proof of Proposition \ref{mpooling_prop}}

Let $x,x' \in \R^N$, and let $\cJ = s(x) \cap s(x')$. 
Suppose first that $\cC_{s(x)} \cap \cC_{s(x')} \neq \emptyset$.
Since $\| P_\infty x - P_\infty \| \geq \|\restr{|\cF_{s} x| - | \cF{s}x' |}{\cJ} \|$, 
it results that
\begin{equation}
\label{couscous1}
d(x,x') A_{s(x),s(x')} \leq \| P_\infty x - P_\infty x' \|
\end{equation}
by Proposition \ref{balanprop} and by definition (\ref{mpool_bobo}).
%
%If $s(x)=s(x')=s$, the two points have the same switches, 
%therefore $P_\infty(x) = | \cF_{s} x | $, $P_\infty(x') = | \cF_{s} x' | $ and hence
%\begin{equation}
%\label{couscous1}
%d(x,x') A_{\cF_{s}} \leq \| P_\infty x - P_\infty x' \|
%\end{equation}
%by Proposition \ref{balanprop}.

Let us now suppose $\cC_{s(x)} \cap \cC_{s(x')}=\emptyset$, 
and let $z = P_\infty x - P_\infty x'$. 
It results that $ z = | \cF_{s(x)} x | - | \cF_{s(x')} x' | \in \R^K$, and hence we can 
split the coordinates $(1\dots K)$ into $\Omega$, $\Omega^c$ such that 
\begin{eqnarray*}
\restr{z}{\Omega} &=& \restr{\cF_{s(x)}}{\Omega}(x) -  \restr{\cF_{s(x')}}{\Omega}(x') ~, \\  
\restr{z}{\Omega^c} &=& \restr{\cF_{s(x)}}{\Omega^c}(x) +  \restr{\cF_{s(x')}}{\Omega^c}(x') ~. \\  
\end{eqnarray*}
We shall concentrate in each restriction independently. 
Since $ \restr{\cF_{s(x')}}{\Omega}(x') \in  \restr{\cF_{s(x')}}{\Omega}(\cC_{s(x')})$, it results that
\begin{eqnarray}
\label{couscous2}
\| \restr{z}{\Omega}  \| &\geq& \inf_{y \in \restr{\cF_{s(x')}}{\Omega} } \| \restr{\cF_{s(x)}}{\Omega}(x) - y \| \nonumber \\
&\geq & \| \restr{\cF_{s(x)}}{\Omega}(x) \| \, \cdot \, |\sin( \beta(s(x),s(x'), \Omega))|  ~.
%& \geq & \lambda_{-}(\restr{\cF_{s(x)}}{\Omega})\, \cdot \,  |\sin( \beta(s(x),s(x'), \Omega))| \, \cdot \, \| x \| ~.
\end{eqnarray}
%HERE UPDATE WITH THE BETTER BOUND FOR THE NORM (WE USE THE FACT THAT WE ARE IN THE RIGHT CONE).
Since by definition $$ \forall~k~,~\sum_{j \in I_k} | \langle x, f_j \rangle |^2 \leq \frac{1}{|I_k|} |\langle x, f_{s(x)_k} \rangle |^2~,$$
it results, assuming without loss of generality that all pools have equal size ( $|I_k| = \frac{M}{K})$, 
\begin{eqnarray}
\forall~x \in \cC_s~,~ \| \restr{\cF_{s(x)}}{\Omega}(x) \| &\geq&\sqrt{\frac{K}{M}}  \| \restr{\cF}{\Omega}(x) \| \nonumber \\
&\geq& \sqrt{\frac{K}{M}}  \lambda_{-}(\restr{\cF}{\Omega})\|x\|~.
\end{eqnarray}
Equivalently, since $ \restr{\cF_{s(x)}}{\Omega}(x) \in  \restr{\cF_{s(x)}}{\Omega}(\cC_{s(x)})$ we also have
\begin{equation}
\label{couscous3}
\| \restr{z}{\Omega}  \|  \geq \sqrt{\frac{K}{M}}  \lambda_{-}(\restr{\cF}{\Omega}) \cdot \, \left |\sin( \beta(s(x),s(x'), \Omega)) \right | \, \cdot \,\| x' \| ~.
\end{equation}
It follows that 
\begin{eqnarray}
\| \restr{z}{\Omega}  \|  &\geq & \sqrt{\frac{K}{M}}  \lambda_{-}(\restr{\cF}{\Omega}) \left |\sin( \beta(s(x),s(x'), \Omega)) \right | \max(\| x \|, \|x'\|) \nonumber \\
&\geq & \sqrt{\frac{K}{4M}} \lambda_{-}(\restr{\cF}{\Omega}) \left |\sin( \beta(s(x),s(x'), \Omega)) \right | d(x,x')~.
\end{eqnarray}
By aggregating the bound for $\Omega$ and $\Omega^c$ we obtain (\ref{cc3})  $\square$.
%Finally, (\ref{cc4}) follows by observing that 
%$$\beta(s,s', \Omega) \geq \alpha(\restr{\cF_s}{\Omega},\restr{\cF_{s'}}{\Omega})$$
%for all $s,\, s'$ and $\Omega$ $\square$.

\subsubsection{Maxout}
These results easily extend to the so-called Maxout operator \cite{maxout}, 
defined as $x \mapsto MO(x)=\{ \max_{j \in I_k} \langle x, f_j \rangle \, ; \, k=1\dots K\}$. 
By redefining the switches of $x$ as
\begin{equation}
\label{switches_maxout}
s(x)=\{ j \, ; \, \langle x, f_j \rangle  > \max( \langle x, f_{j'}\rangle \,; \,\forall~j' \in \mbox{pool}(j)\} ~,
\end{equation}
the following corollary computes a Lower Lipschitz bound of $MO(x)$:
\begin{corollary}
\label{cormaxout}
The Maxout operator $MO$ satisfies (\ref{cc34}) with $A(s,s')$ defined 
using the switches (\ref{switches_maxout}).
\end{corollary}

\subsubsection{$\ell_1$ Pooling}
\label{sec:l1sup}
Propostion \ref{mpooling_prop} can be used to 
obtain a bound of the lower Lipschitz constant 
of the $\ell_1$ pooling operator. 

Observe that for $x \in \mathbb{R}^n$,  
$$\| x \|_1 = \sum_i |x_i| = \max_{\epsilon_i = \pm 1} | \langle x, \epsilon \rangle |~.$$
It results that $P_1(x; \mathcal{F}) \equiv P_\infty(x; \widetilde{\mathcal{F}})$, with 
$$\widetilde{\mathcal{F}} = ( \tilde{f}_{k,\epsilon}= \sum_i \epsilon(i) f_{k,i} \,; \, k=1\dots,K \,; \epsilon \in \{-1,1\}^L \}~.$$
Each pool $\widetilde{\cF}_k$ can be rewritten as $\widetilde{\cF}_k = H_L \cF_k$, where 
$H_L$ is the $L \times 2^L$ Hadamard matrix whose rows contain the $\epsilon$ vectors. 
One can verify that $H_L^T H_L = 2^L {\bf 1}$, which implies that 
for any $\Omega \subseteq \{1\dots K\}$, 
$\lambda_{-}(\restr{\widetilde{\cF}}{\Omega}) = 2^{L/2} \lambda_{-} (\restr{{\cF}}{\Omega}) $.
It results that
\begin{corollary}
\label{l1corol}
The $\ell_1$ pooling operator $P_1$ satisfies
\begin{equation}
\label{cc3}
\forall\, x,x'~,~d(x,x') \left(\min_{s,s'} \tilde{A}(s,s') \right) \leq \| P_1(x) - P_1(x') \|~,
\end{equation}
where $d(x,x')=\min(\|x-x'\|,\|x+x'\|)$ and 
\begin{eqnarray*}
%\label{mpool_bobo}
\tilde{A}(s,s') &=& \max \Big \{ \min_{\Omega \subseteq \cJ(s,s')} \sqrt{\lambda_{-}^2(\widetilde{\cF}_\Omega) + \lambda_{-}^2(\widetilde{\cF}_{\cJ - \Omega})} ~, \nonumber \\
& & \frac{1}{2} \, \min_{\Omega \subseteq \{1\dots K\}} \sqrt{ 
\Lambda_{s,s',\Omega}^2+ \Lambda_{s,s',\Omega^c}^2  }\Big \}  ~,
\end{eqnarray*}
with $s,s'$ and $\beta(s,s')$ are defined on the frame $\widetilde{\mathcal{F}}$.
\end{corollary}
Similarly as in Corollary \ref{mpoolrectprop}, one can obtain 
a similar bound for the Rectified $\ell_1$ pooling. 

%Rectified L1 case 
%\begin{proposition}
%
%\end{proposition}

%$\ell_2$ is the limit case: 
%$\| x \|_2 = \sup_{\|e \|=1} \langle x, e \rangle$.
%Does the bound degrade? How does it compare to the previous bound?

\subsection{Proof of Corollaries \ref{mpoolrectprop} and \ref{cormaxout} }
The result follows immediately from Proposition \ref{mpooling_prop}, 
by replacing the phaseless invertibility condition of Propostion \ref{balanprop}
by the one in Proposition \ref{halfrect_inj}. $\square$.

\subsection{Proof of Proposition \ref{randomprop}}
 Proposition \ref{randomprop} also extends to the maxout case.  We restate it here with the extra result:
\begin{proposition}
\label{randomprop_sup}
Let $\cF=(f_1,\dots,f_M)$ be a random frame of $\R^N$, organized
into $K$ disjoint pools of dimension $L$.
 Then these statements hold with probability $1$:
\begin{enumerate}
\item $P_p$ is injective (modulo $x \sim -x$) if $K \geq 4N$ for $p=1,\infty$, 
and if $K \geq 2N-1$ for $p=2$.
\item The Maxout operator $MO$ is injective if $K \geq 2N+1$.
\end{enumerate}
\end{proposition}
Let us first prove (i), with $p=\infty$. 
Let $x, \, x' \in \R^N$ such that $P_\infty(x) = P_\infty(x')$, 
and let $s=s(x)$, $s'=s(x')$. 
The set of $K$ pooling measurements is divided into 
the intersection $\cJ(s,s')=\{ k \, ; \, s(x)_k = s(x')_k \}$ 
and its complement $\cJ(s,s')^c = \{ k \, ; \, s(x)_k \neq s(x')_k \}$.
Suppose first that $|\cJ(s,s')| \geq 2N-1$. Then 
it results 
that we can pick $d=\lceil \frac{|\cJ(s,s')|}{2} \rceil \geq N$ elements of $\cJ(s,s')$ 
to form a frame $V$, such that either $x-x' \in \mbox{Ker}(V)$ 
or $x+x \in \mbox{Ker}(V)$. Since a random frame of dimension $\geq N$ spans
$\R^N$ with probability $1$, it results that $x=\pm x'$.
Suppose otherwise that $|\cJ(s,s')| < 2N-1$. It follows that 
$|\cJ(s,s')^c| \geq 2N+1$, and hence that any partition of 
$\cJ(s,s')^c$ into two frames will contain always a frame $\restr{\cF}{\Omega}$ with at least $N+1$ columns. 
Since two random subspaces of dimension $N$ in $\R^{M}$ have nonzero largest principal 
angle with probability $1$ as long as $K>N$, it results that $\Lambda_{s,s',\Omega} > 0$
and hence that $\mbox{Prob}(|\restr{P_\infty(x)}{\cJ(s,s')^c}| = |\restr{P_\infty(x')}{\cJ(s,s')^c}|)=0$.
The case $p=1$ is proved identically thanks to Corollary \ref{l1corol}.

Finally, in order to prove (ii) we follow the same strategy. If $|\cJ(s,s')| \geq N$, then 
$MO(x) = MO(x') \Rightarrow \, x=x'$ with probability $1$ since $\restr{\cF}{\cJ}$ spans $\R^N$ with probability $1$. 
Otherwise it results that $|\cJ(s,s')^c|  \geq N+1$, which implies $MO(x)\neq MO(x')$, since 
two random subspaces of dimension $N$ in $\R^{|\cJ(s,s')^c|}$ have $0$ intersection with probability $1$. 

Let us now prove the case $p=2$.
We start drawing a random basis for each of the pools $F_1, \dots, F_K$.
From proposition \ref{p2lips}, it follows that we have to check that if $M\geq 2N$, the quantity
$$\min_{F' = U\, F\, ,\, U^T U = {\bf 1}} \min_{\Omega \subseteq \{1\dots M \} } \lambda_{-}^2( F'_\Omega) + \lambda_{-}^2(F'_Omega^c) > 0$$
with probability $1$.
If $M\geq 2N-1$, it follows that either $\Omega$ has the property that it 
intersects at least $N$ pools, either $\Omega^c$ intersects $N$ pools.
Say it is $\Omega$. Now, for each pool with nonzero intersection, say $F_k$, we have that
$$\| (F'_k)^T y \| \geq \frac{1}{\sqrt(L)} |\langle  f_{k,j}, y \rangle| $$
for some $f_{k,j}$ belonging to the initial random basis of $F_k$.
It results that
$$\lambda_{-}^2(F'_\Omega) \geq \frac{1}{\sqrt(L)} \lambda_{-}^2(F^*) ~,$$
where $F^*$ is a subset of $N$ columns of the original frame $\cF$, which means
$$\lambda_{-}^2(F'_\Omega) \geq \frac{1}{\sqrt(L)} \lambda_{-}^2(F*) > 0~.$$ 
$\square$.

\section{Notes on changes from cycle 1}
The mathematical results have been essentially rewritten, for clarity as well as to sharpen the bounds.  The proofs are now in the supplementary material, as requested by the reviewers.  We have used the extra space to expand the indroduction, conclusion, and intro to the experiments, in part to  to explain the connections between the theoretical and experimental parts of the paper, as requested by the reviewers.  We also added results on the invertibility of random modules.

We have edited the text in the experiments section and in the captions of the figures to clarify them.  Each curve is described in the caption and the text; the graphs are also now specifically referenced in the analysis bullets in section \ref{sec:experimental_analysis}.

The introduction and conclusion more explicitly address take messages.  Note that the take home message is not of the form ``this is how to design a network'', but rather, ``these conditions allow (stable) inversion''.    We are sympathetic to the reviewers desire for a take home message giving insight into the actual design of networks for practical applications.  That is, of course, the ultimate goal of a mathematical analysis of a learning algorithm.  However, if the standard for theoretical papers analyzing deep models is that they lead immediately to design suggestions with associated performance increases on benchmarks, it is unlikely that there will ever be a mature enough theory to give honest design suggestions.  

Finally, we reprint larger versions of the figures below.

\begin{figure*}
\begin{center}
\includegraphics[width=.49\linewidth,height=.3\linewidth]{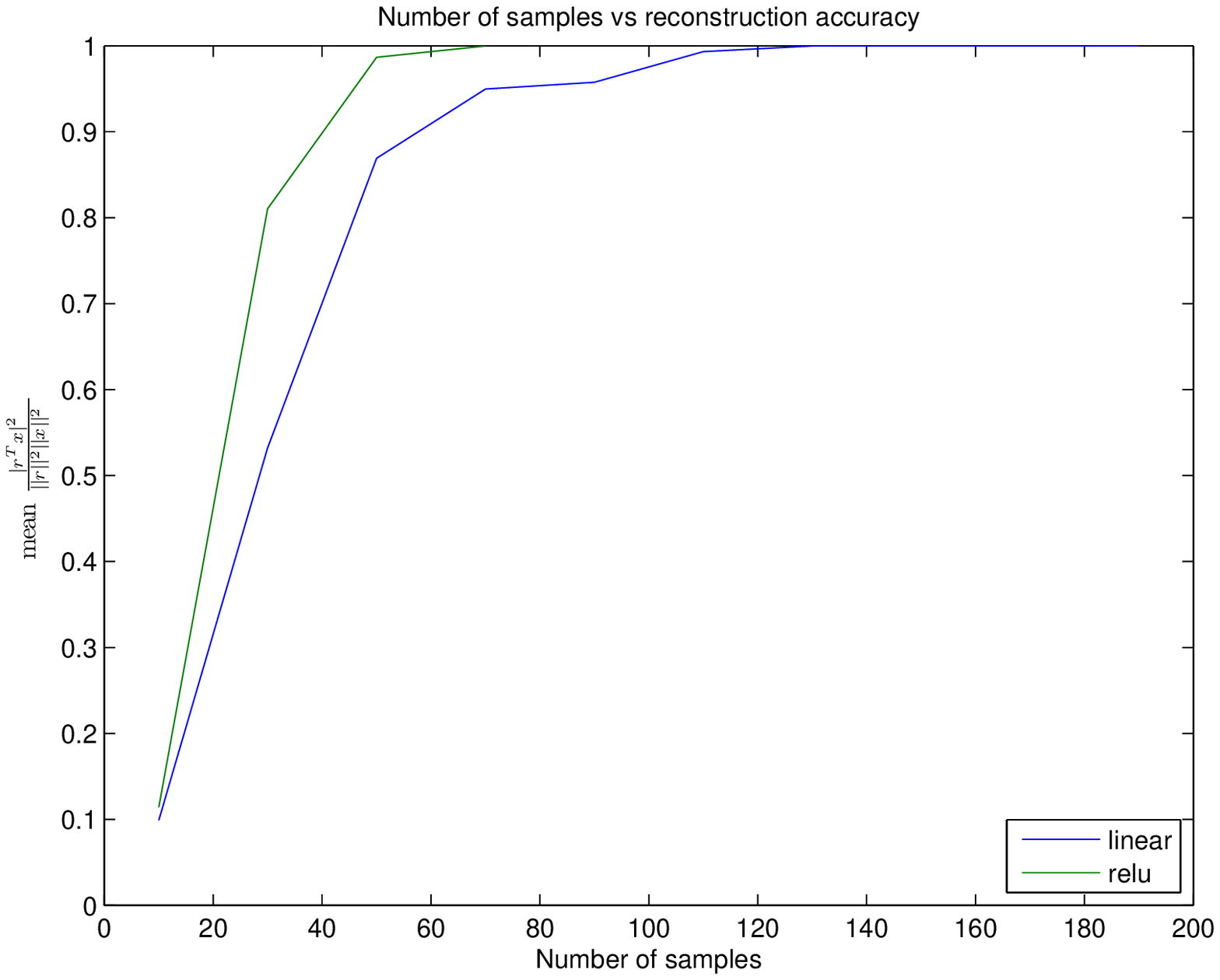} 
\includegraphics[width=.49\linewidth,height=.3\linewidth]{lp_d40_fullrandom_l1.eps} 
\includegraphics[width=.49\linewidth,height=.3\linewidth]{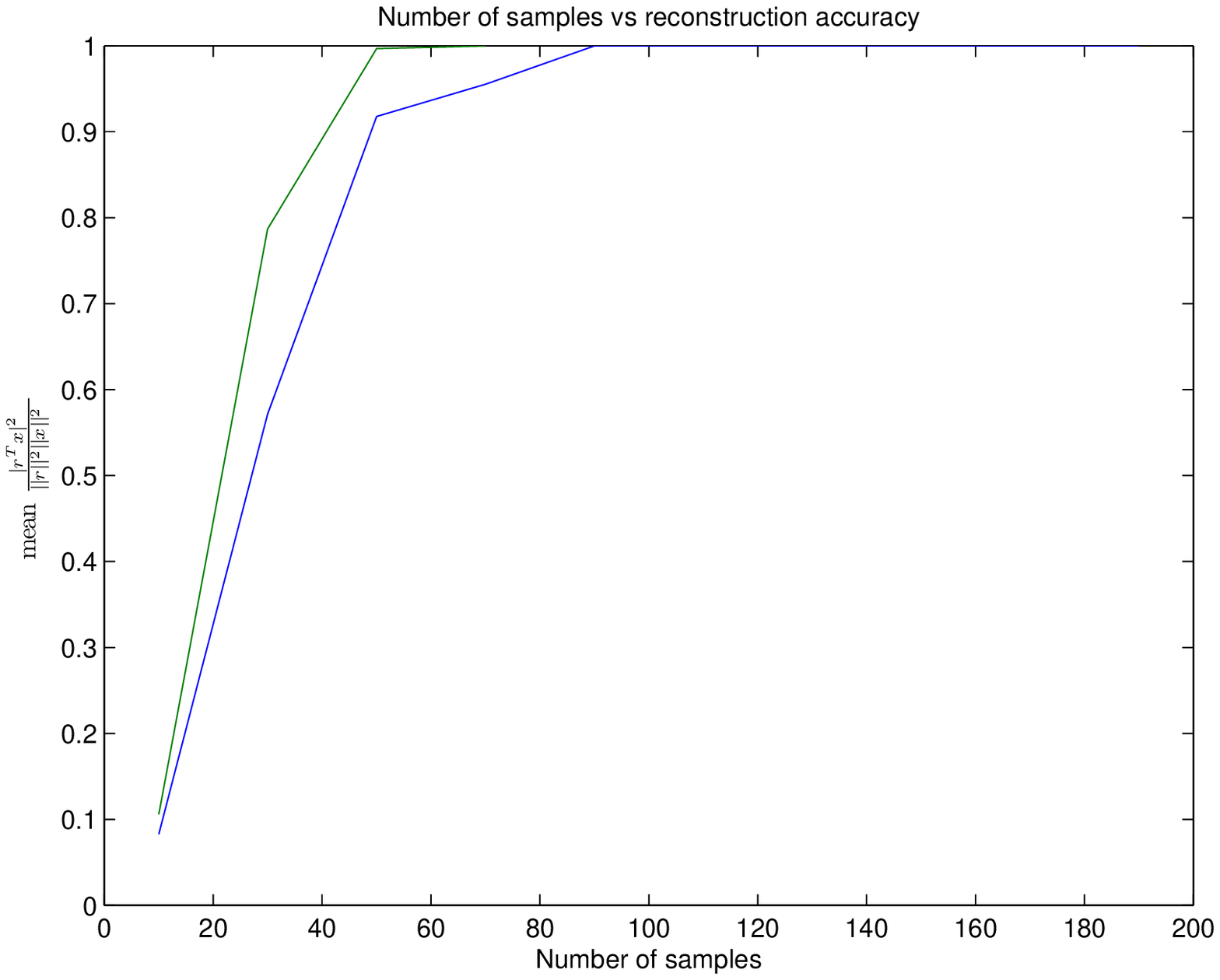}
\includegraphics[width=.49\linewidth,height=.3\linewidth]{lp_d40_fullrandom_l2.eps}
\includegraphics[width=.49\linewidth,height=.3\linewidth]{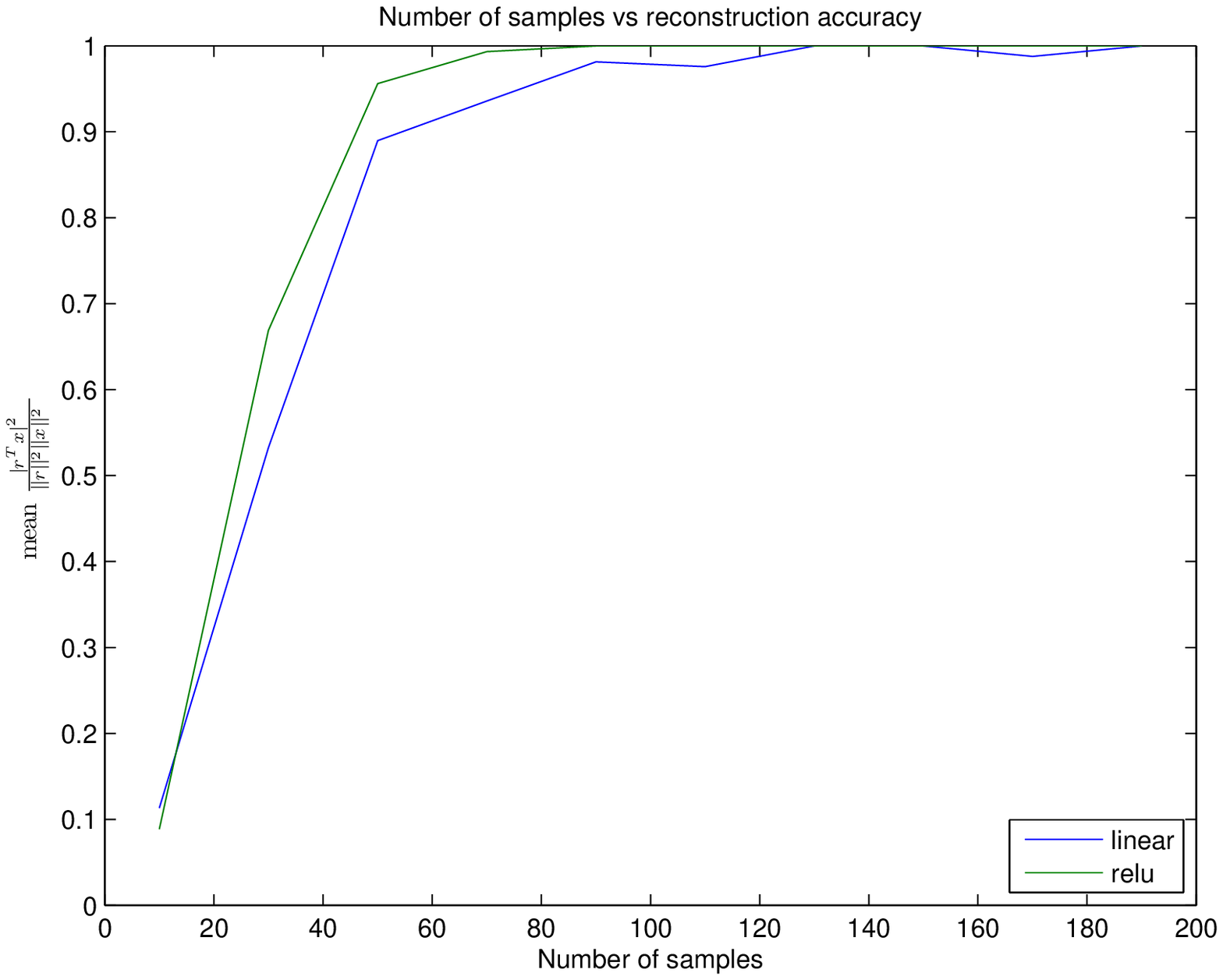}
\includegraphics[width=.49\linewidth,height=.3\linewidth]{lp_d40_fullrandom_linfty.eps}
\end{center}
\caption{Average recovery angle using alternating projections on random data.   The vertical axis measures the average value of $|r^Tx|^2/(||r||^2||x||^2)$ over 50 random test points.  The horizontal axis is the number of measurements (the size $m$ of the analysis dictionary is twice the $x$ axis in this experiment).  The top row is $\ell_1$ pooling, the middle $\ell_2$, and the bottom max pooling.  In the left column each $x$ is Gaussian i.i.d. in $\R^{20}$, on the right, in $\R^{40}$.  The dark blue curve is alternating minimization, and the green curve is alternating minimization with half rectification; both with random initialization.\label{f:random_random_sup}}
\end{figure*}
\label{mnist_experiments}
\begin{figure*}
\begin{center}
\includegraphics[width=.49\linewidth,height=.3\linewidth]{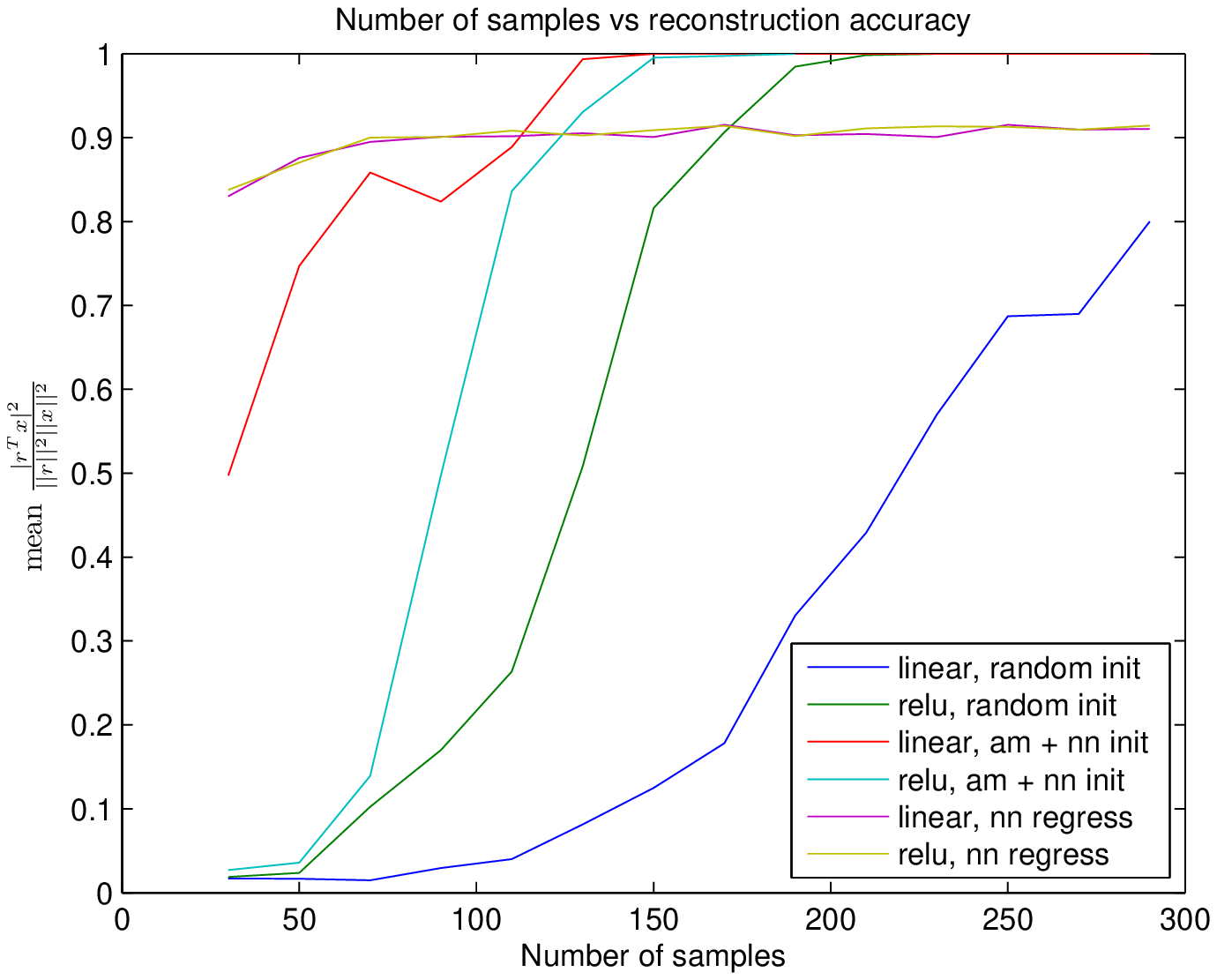} 
\includegraphics[width=.49\linewidth,height=.3\linewidth]{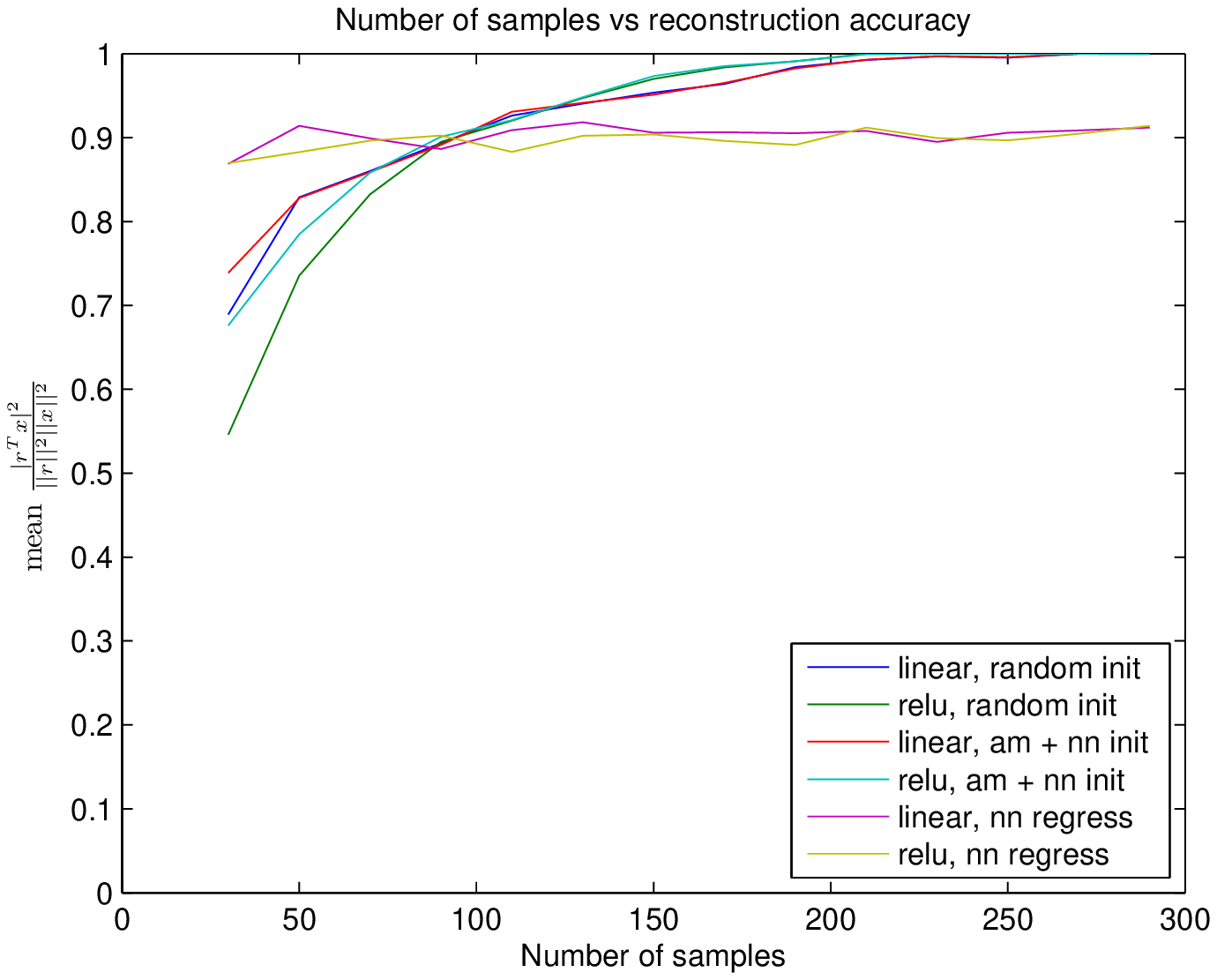} 
\includegraphics[width=.49\linewidth,height=.3\linewidth]{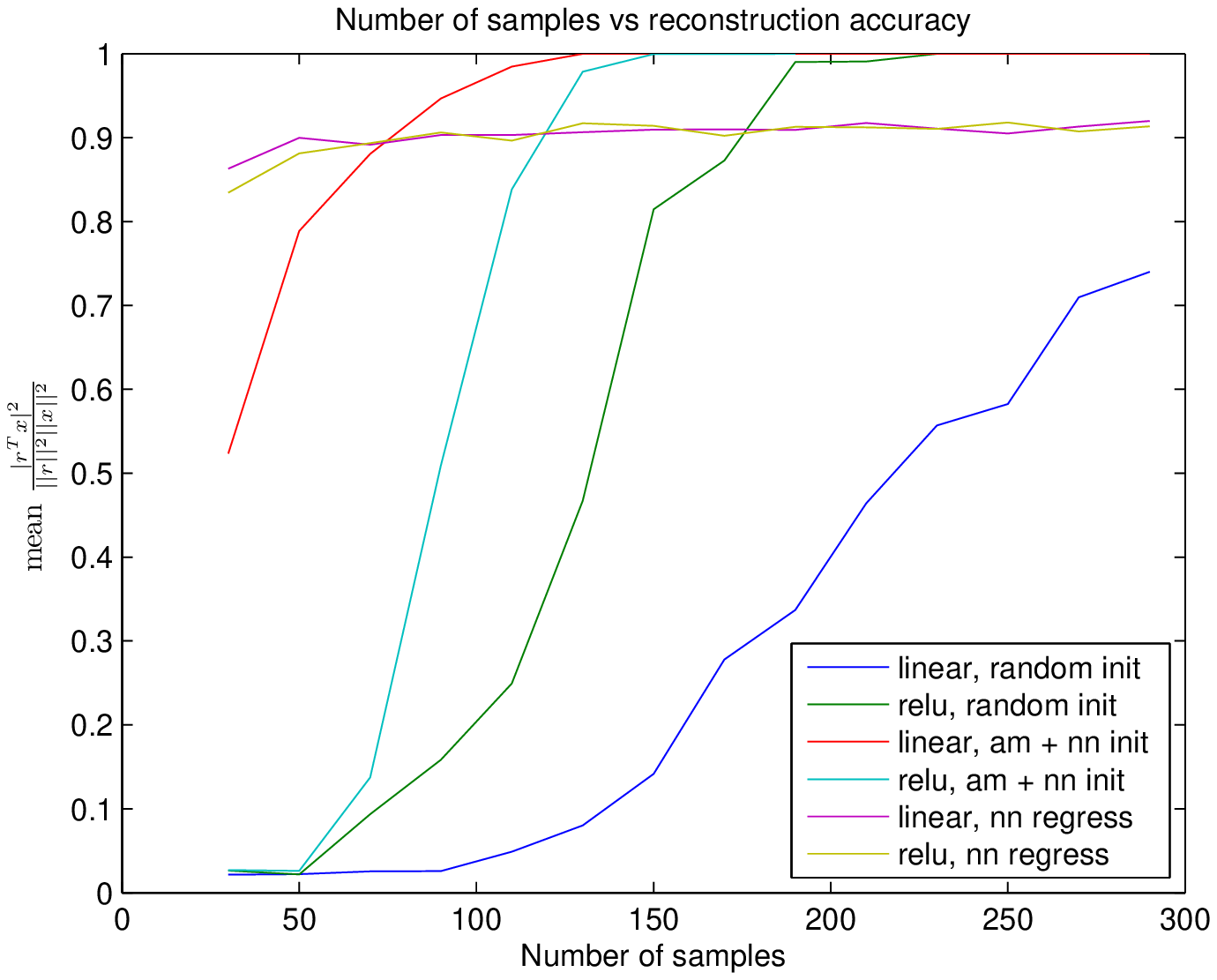}
\includegraphics[width=.49\linewidth,height=.3\linewidth]{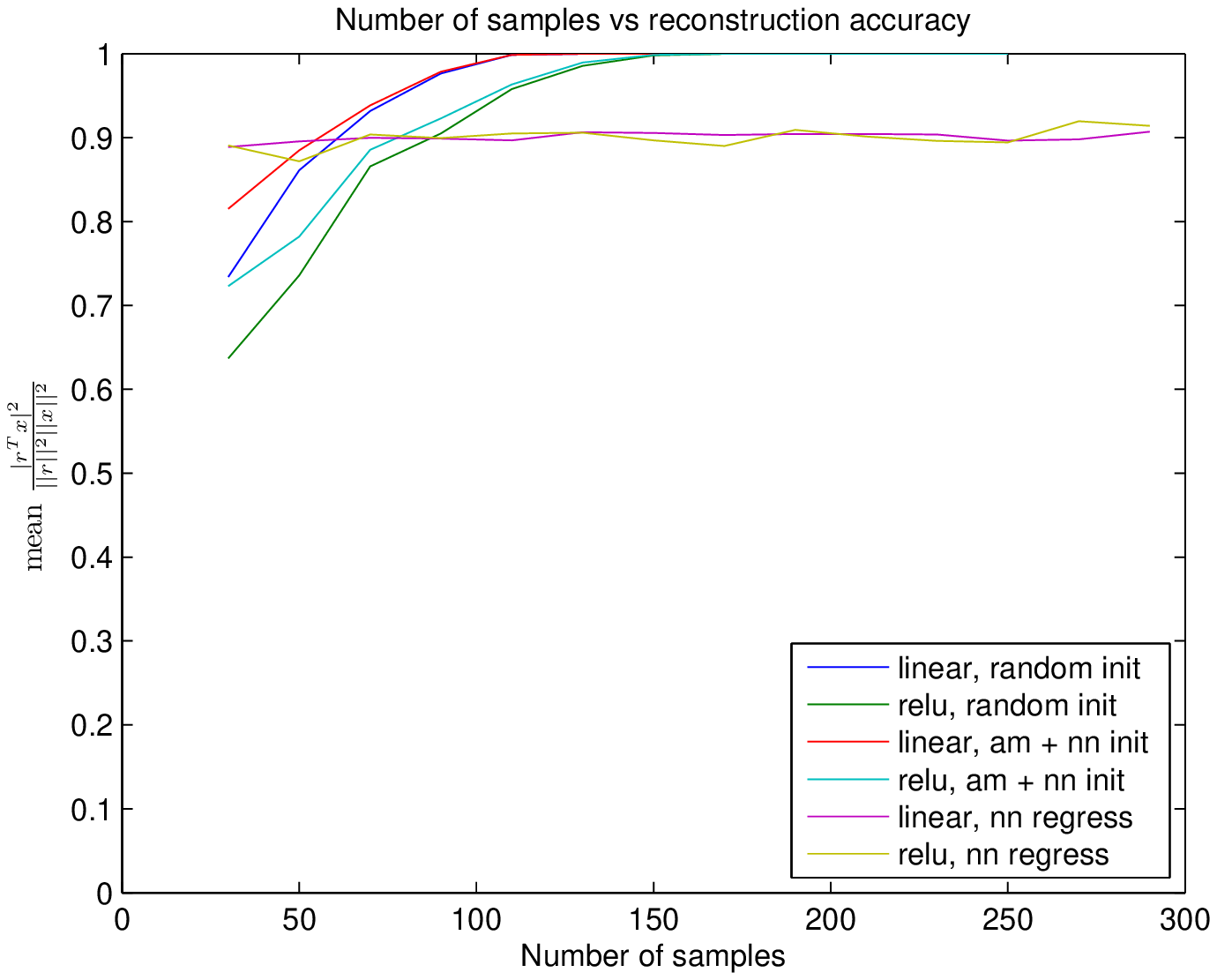}
 \includegraphics[width=.49\linewidth,height=.3\linewidth]{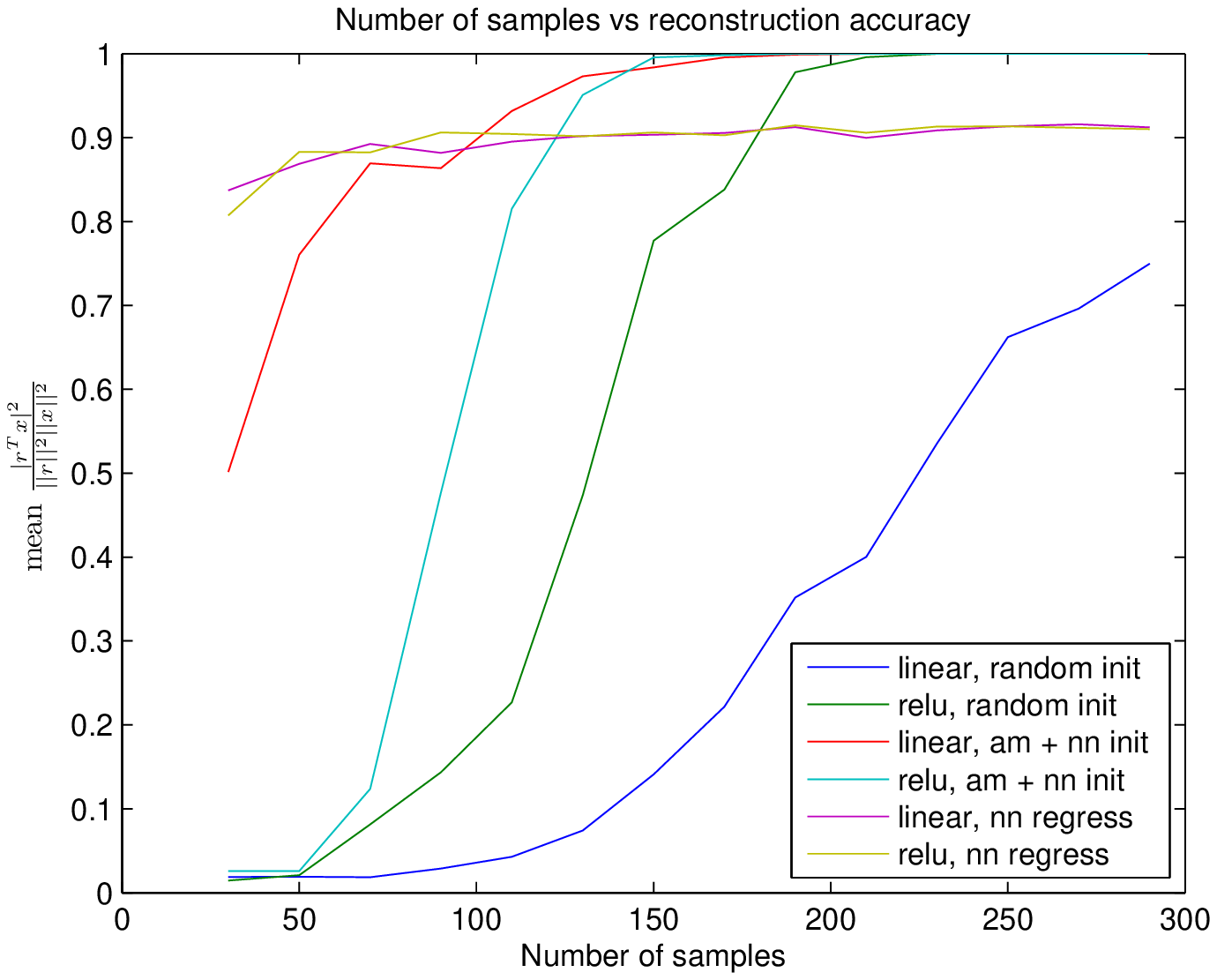}
 \includegraphics[width=.49\linewidth,height=.3\linewidth]{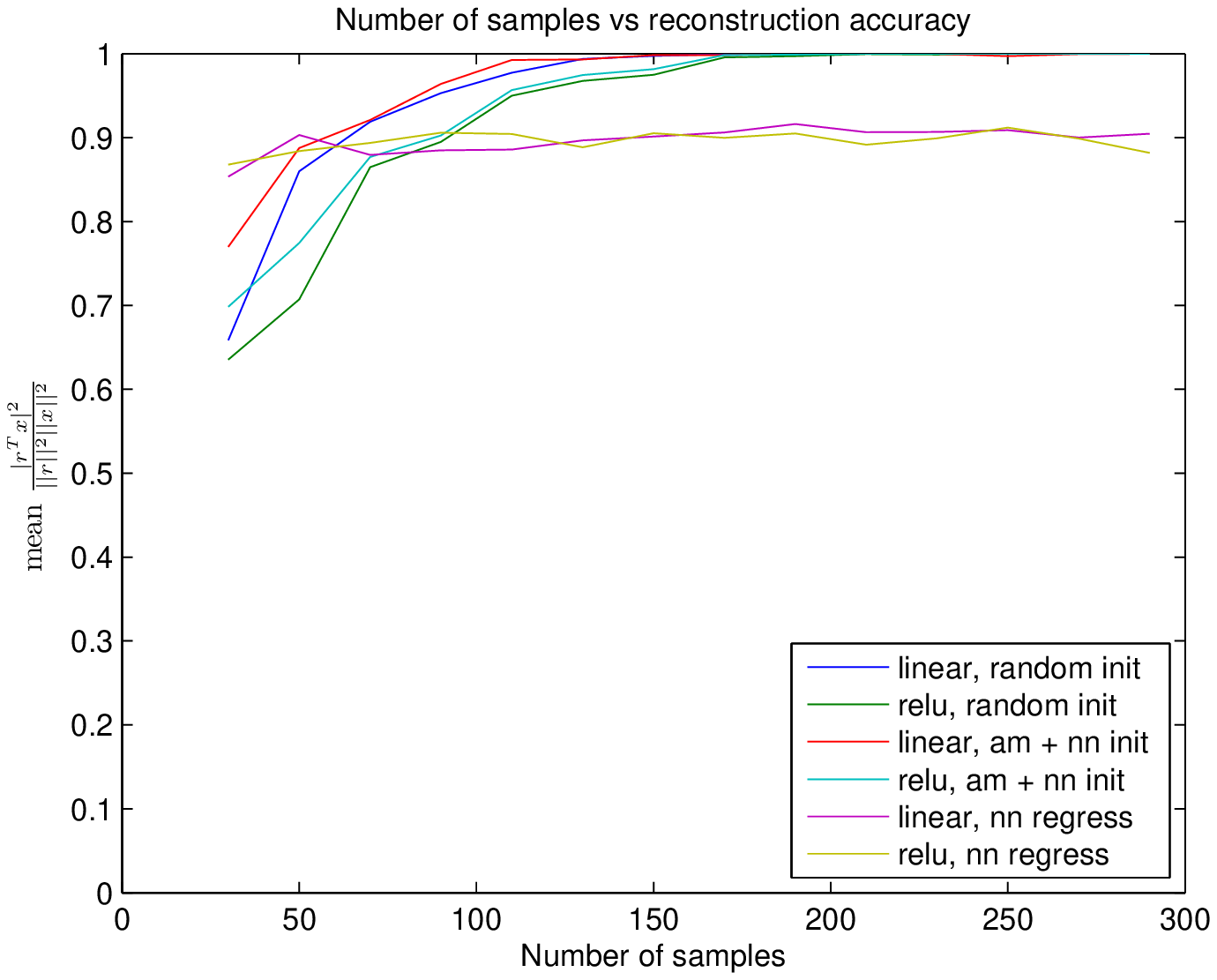}
\end{center}
\caption{Average recovery angle using alternating projections on MNIST data points.   The vertical axis measures the average value of $|r^Tx|^2/(||r||^2||x||^2)$ over 50 random test points.  The horizontal axis is the number of measurements (the size of the analysis dictionary is twice the $x$ axis in this experiment).  The top row is $\ell_1$ pooling, the middle $\ell_2$, and the bottom max pooling.  In the left column the analysis dictionary is Gaussian i.i.d.;  in the right column, generated by block OMP/KSVD with $5$ nozero blocks of size 2.  The dark blue curve is alternating minimization, and the green curve is alternating minimization with half rectification; both with random initialization.  The magenta and yellow curves are the nearest neighbor regressor described in \ref{sec:nn_regress} without and with rectification ; and the red and aqua curves are alternating minimization initialized via neighbor regression, without and with rectification.\label{f:mnist_sup}}
\end{figure*}
\begin{figure*}
\begin{center}
\includegraphics[width=.49\linewidth,height=.3\linewidth]{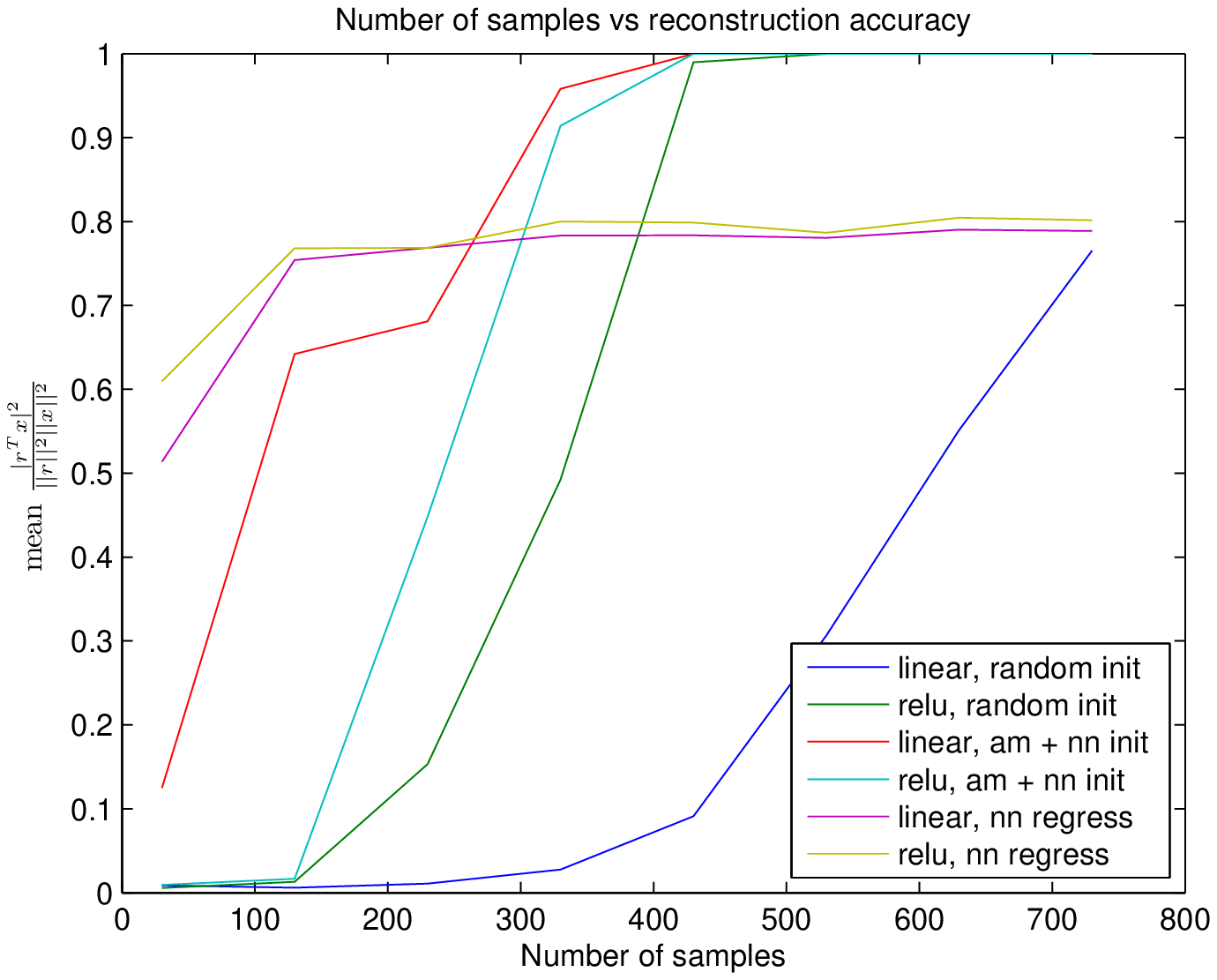} 
 \includegraphics[width=.49\linewidth,height=.3\linewidth]{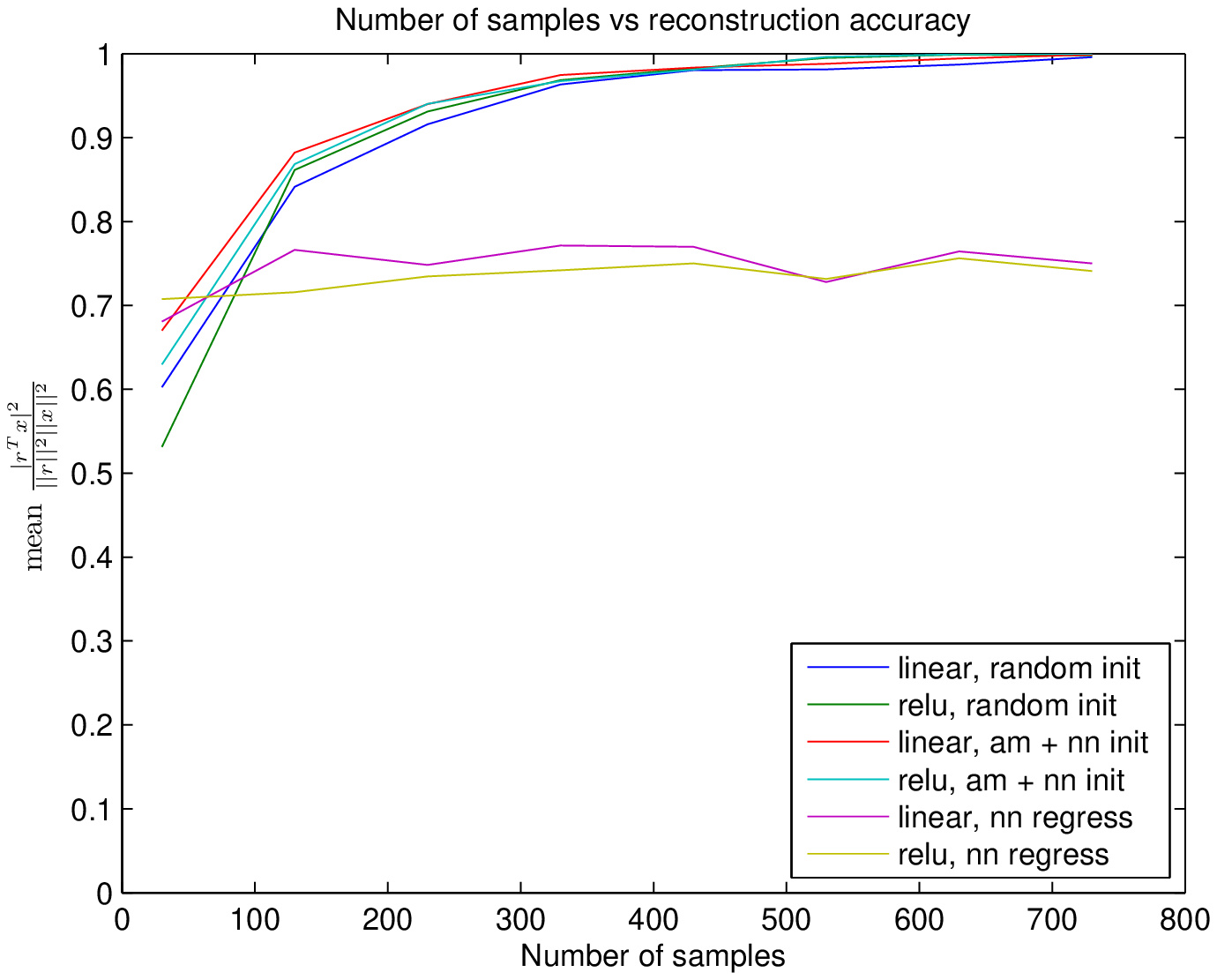} 
\includegraphics[width=.49\linewidth,height=.3\linewidth]{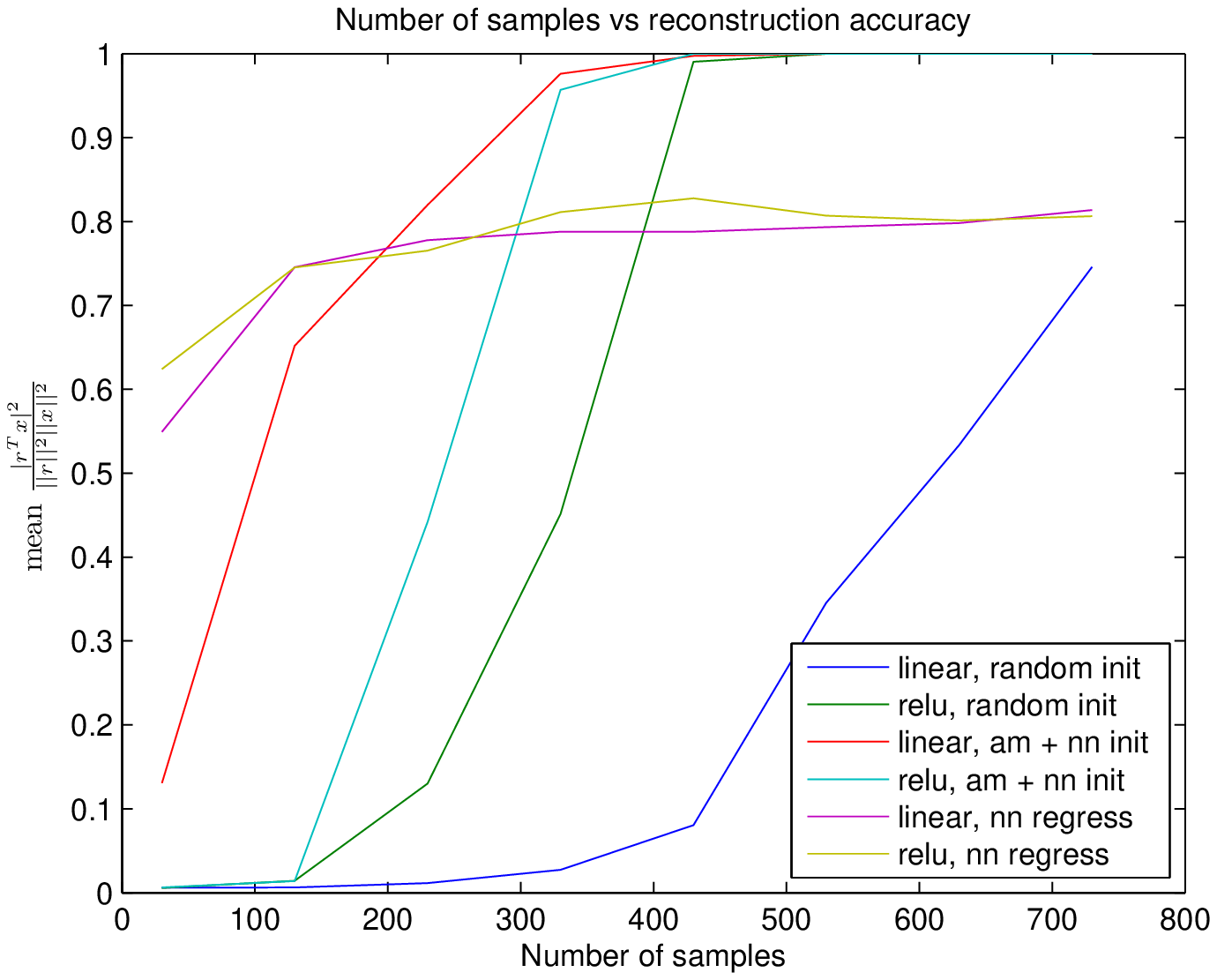}
\includegraphics[width=.49\linewidth,height=.3\linewidth]{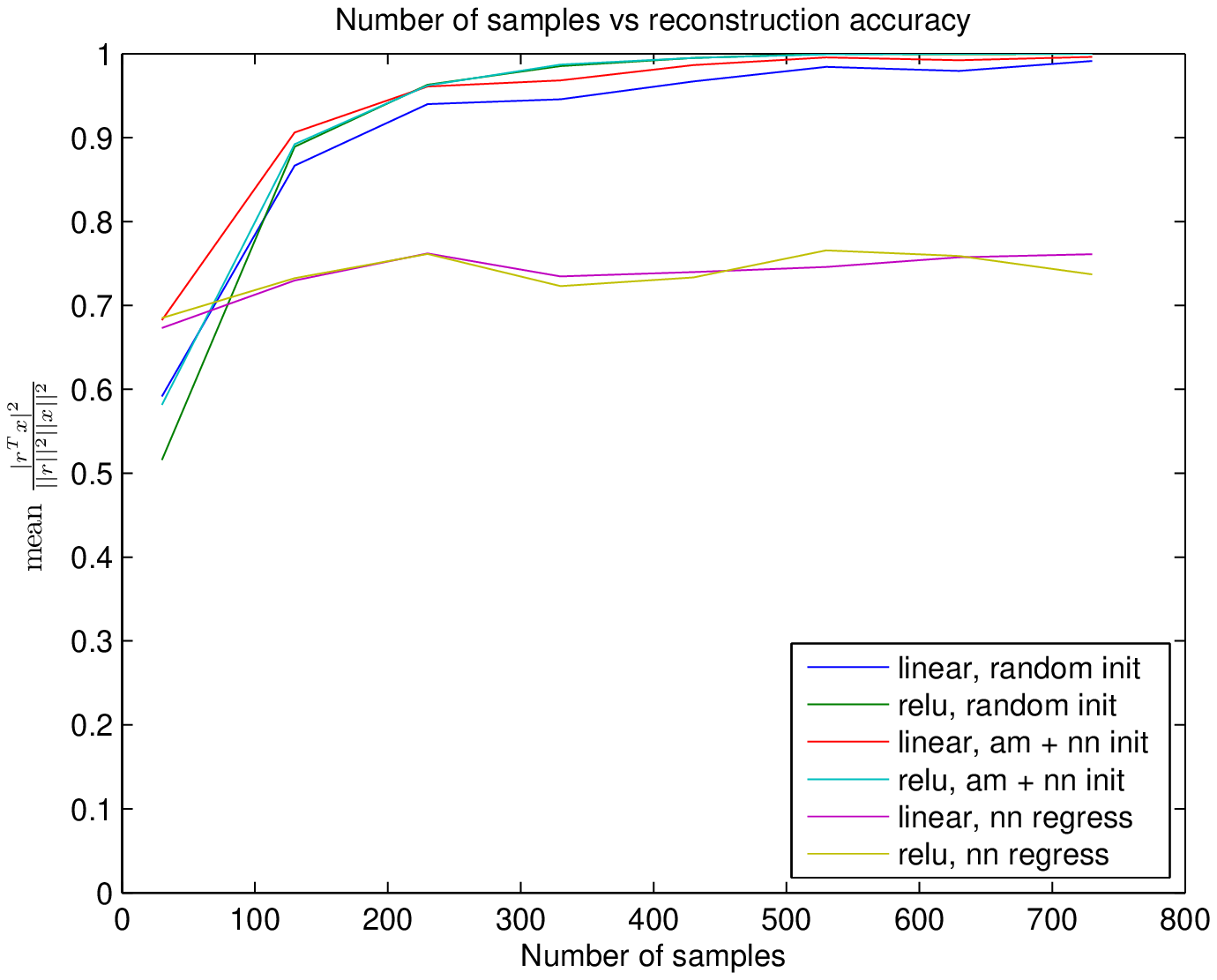}
 \includegraphics[width=.49\linewidth,height=.3\linewidth]{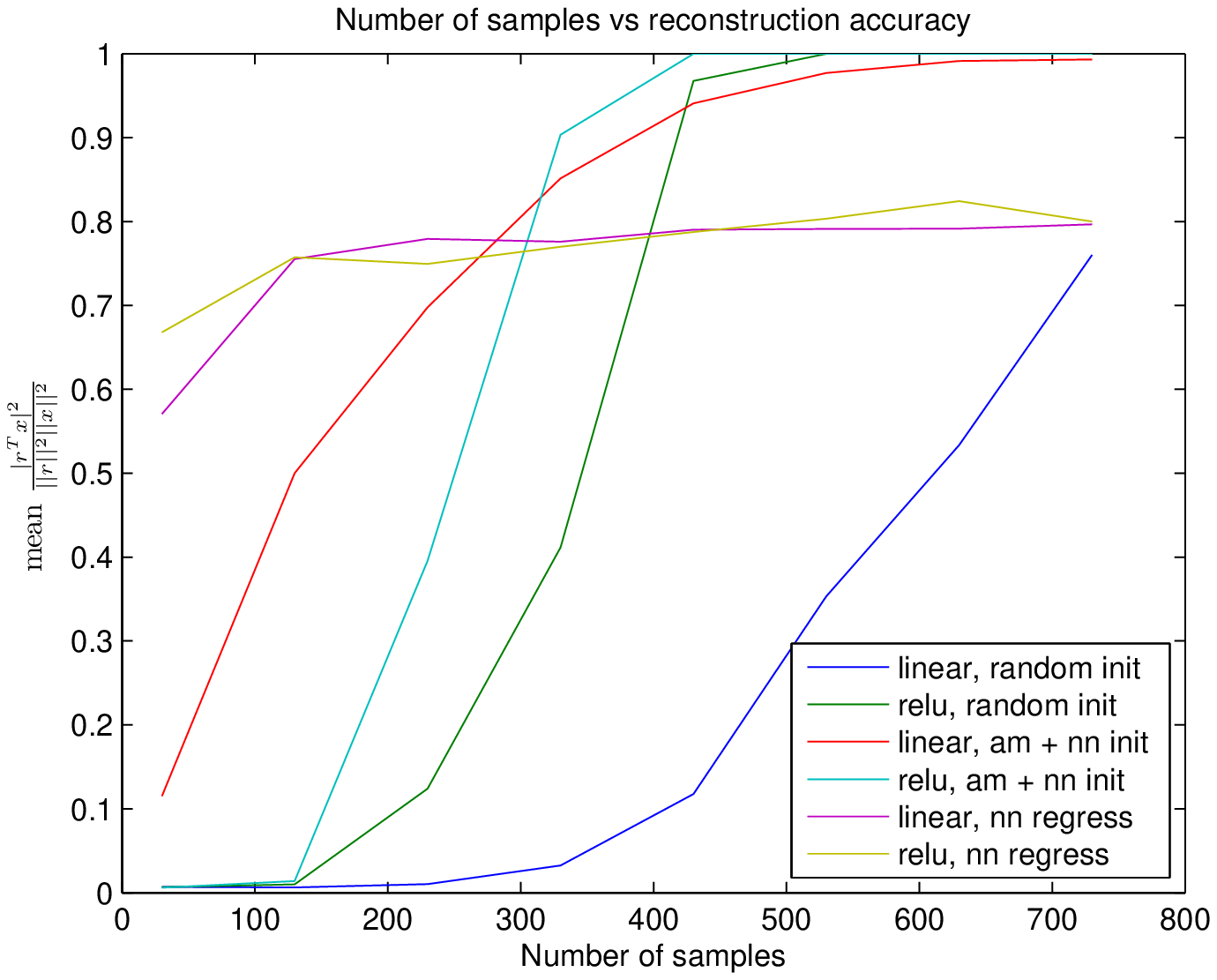}
 \includegraphics[width=.49\linewidth,height=.3\linewidth]{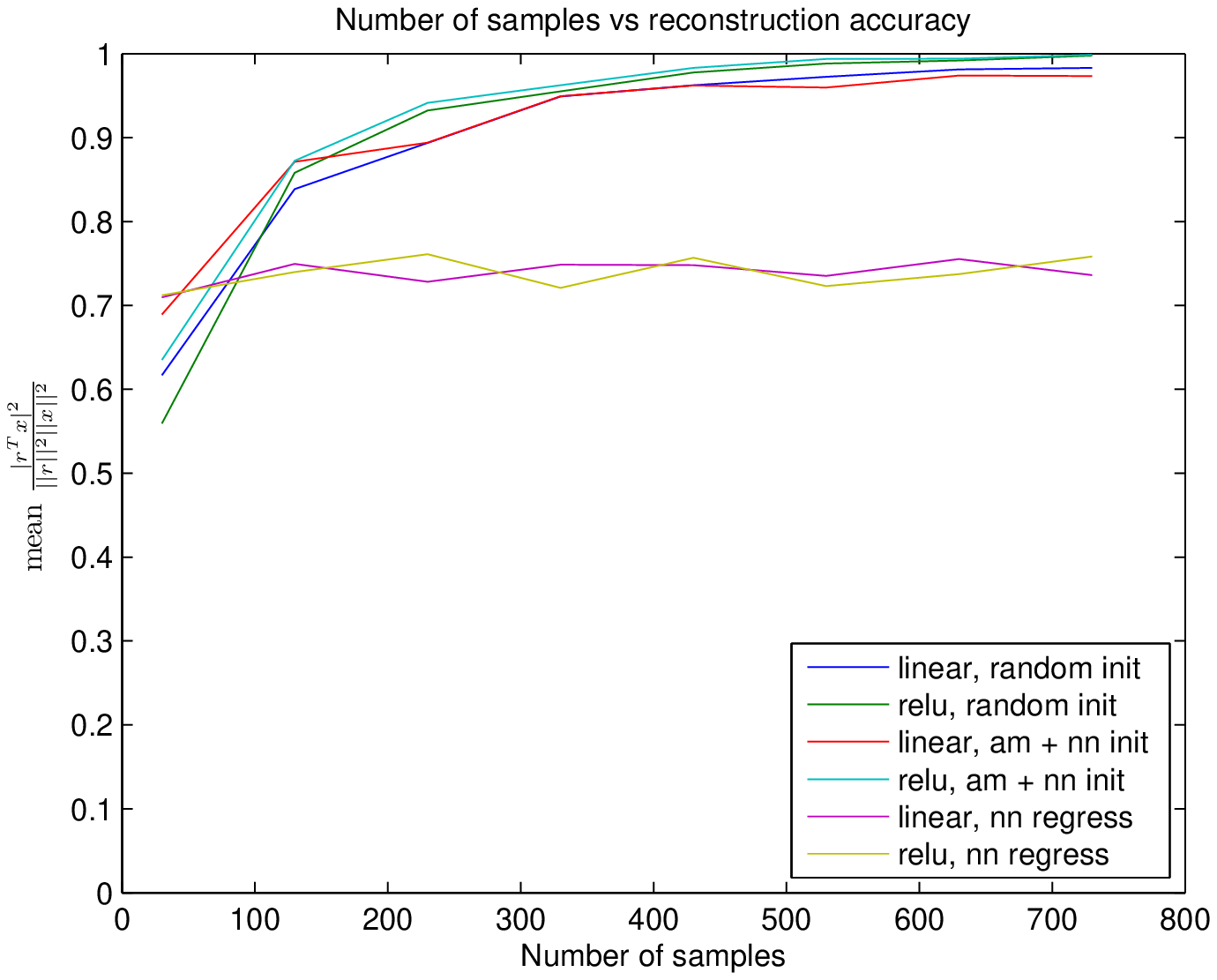}
 \end{center}
\caption{Average recovery angle using alternating projections on image patch data points.   The vertical axis measures the average value of $|r^Tx|^2/(||r||^2||x||^2)$ over 50 random test points.  The horizontal axis is the number of measurements (the size of the analysis dictionary is twice the $x$ axis in this experiment).  The top row is $\ell_1$ pooling, the middle $\ell_2$, and the bottom max pooling.  In the left column the analysis dictionary is Gaussian i.i.d.;  in the right column, generated by block OMP/KSVD with $5$ nozero blocks of size 2. The dark blue curve is alternating minimization, and the green curve is alternating minimization with half rectification; both with random initialization.  The magenta and yellow curves are the nearest neighbor regressor  described in \ref{sec:nn_regress} without and with rectification; and the red and aqua curves are alternating minimization initialized via neighbor regression, without and with rectification.\label{f:patches_sup}}
\end{figure*}

\end{document}